\documentclass[10pt,twocolumn,letterpaper]{article}

\usepackage[pagenumbers]{cvpr} 
\usepackage{fontawesome}
\usepackage[colorlinks=true, linkcolor=black, citecolor=blue, urlcolor=blue]{hyperref}

\usepackage{algorithm}
\usepackage{algpseudocode}

\definecolor{cvprblue}{rgb}{0.21,0.49,0.74}
\usepackage{xcolor}
\usepackage{amsmath}
\usepackage{bm}
\usepackage{color, colortbl}
\definecolor{front-color}{HTML}{F1FFFF}
\definecolor{Gray}{gray}{0.90}
\usepackage{arydshln}
\usepackage{comment}
\usepackage{array}
\usepackage{booktabs}
\usepackage{stfloats}
\fnbelowfloat
\usepackage{lipsum}

\def\paperID{} 
\def\confName{CVPR}
\def\confYear{2026}

\newcommand{\dashedmidrule}{\noalign{\vskip\aboverulesep}\hdashline\noalign{\vskip\belowrulesep}}

\title{Mobile-O: Unified Multimodal Understanding and Generation on Mobile Device}

\author{
    Abdelrahman Shaker$^{1,*,\dagger}$ \quad 
    Ahmed Heakl$^{1,\dagger}$ \quad 
    Jaseel Muhammad$^{1}$ \quad \\
    Ritesh Thawkar$^{1}$ \quad 
    Omkar Thawakar$^{1}$ \quad 
    Senmao Li$^{1}$ \quad 
    Hisham Cholakkal$^{1}$ \quad \\
    Ian Reid$^{1}$ \quad 
    Eric P. Xing$^{1,2}$ \quad 
    Salman Khan$^{1}$ \quad 
    Fahad Shahbaz Khan$^{1,3}$ \\
    \vspace{1pt} \\
    $^{1}$Mohamed bin Zayed University of Artificial Intelligence \quad \\
    $^{2}$Carnegie Mellon University \quad $^{3}$Linköping University \\
    \vspace{5pt} \\
    {\small \raisebox{-2pt}{\includegraphics[height=10pt]{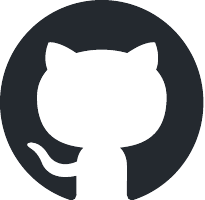}}\hspace{4pt}Codebase: \href{https://github.com/Amshaker/Mobile-O}{\texttt{github.com/Amshaker/Mobile-O}}} \\[1pt]
    {\small \raisebox{-1pt}{\includegraphics[height=10pt]{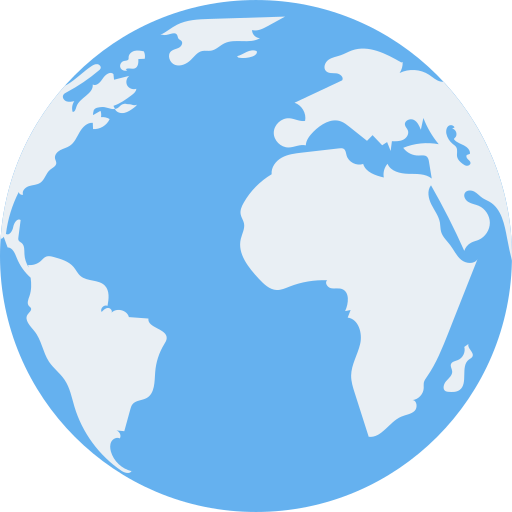}}\hspace{4pt}Project Page: \href{https://amshaker.github.io/Mobile-O}{\texttt{amshaker.github.io/Mobile-O}}} \\[1pt]
    {\small \raisebox{-2pt}{\includegraphics[height=10pt]{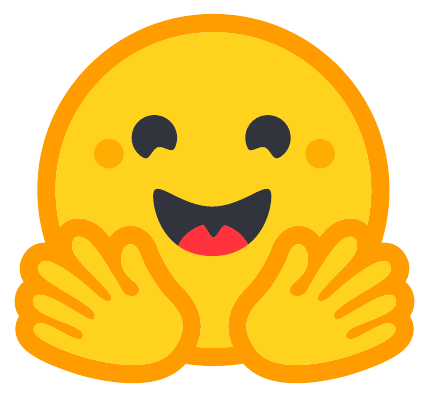}}\hspace{4pt}Models: \href{https://huggingface.co/collections/Amshaker/mobile-o-models}{\texttt{https://huggingface.co/collections/Amshaker/mobile-o-models}}} \\[1pt]
    {\small \raisebox{-2pt}{\includegraphics[height=10pt]{assets/hf-logo.pdf}}\hspace{4pt}Datasets: \href{https://huggingface.co/collections/Amshaker/mobile-o-datasets}{\texttt{https://huggingface.co/collections/Amshaker/mobile-o-datasets}}}
}

\begin{document}

\teaser{
\begin{center}
    \includegraphics[width=0.99\textwidth]{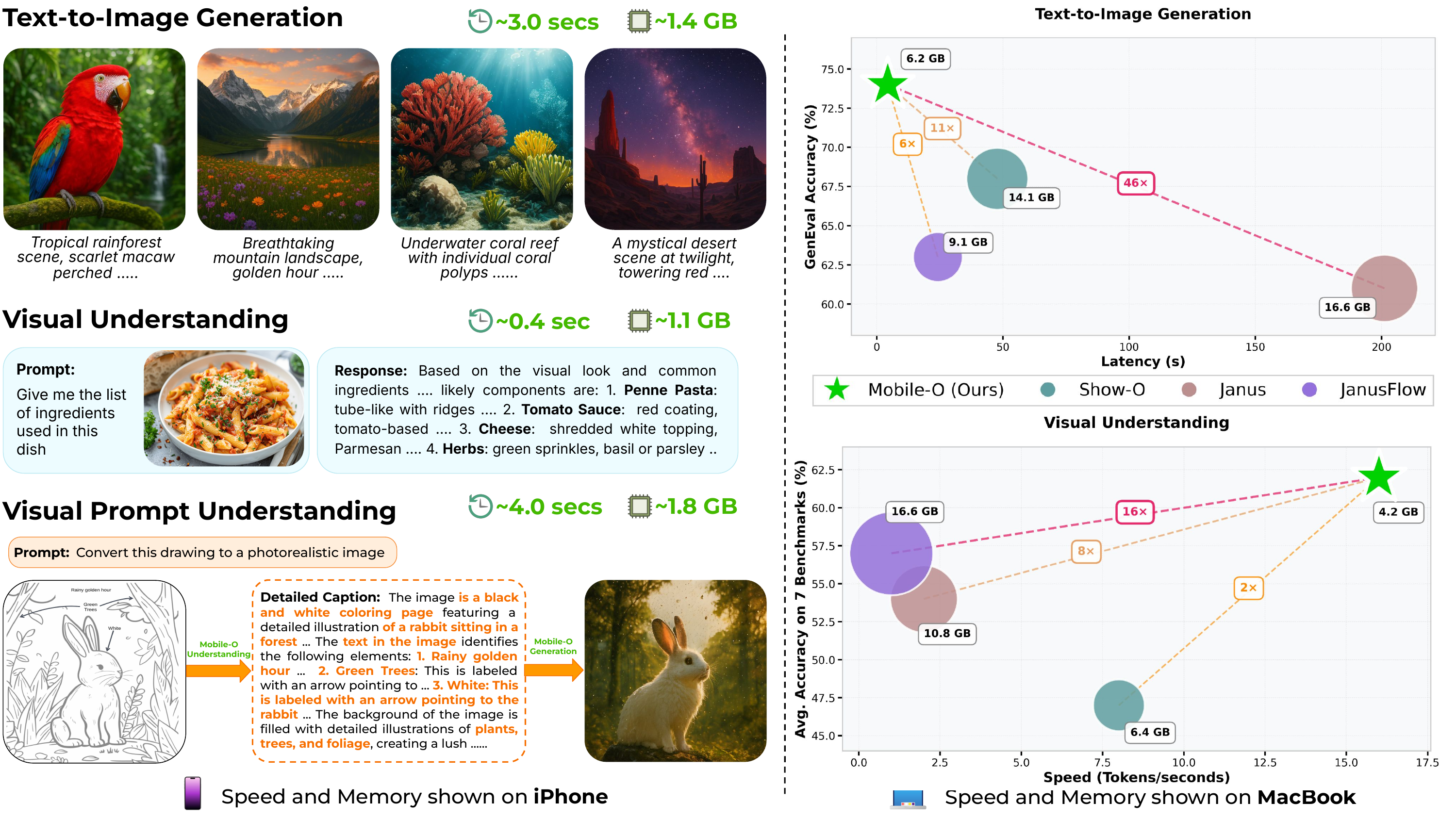}
    \captionof{figure}{
        \textbf{Comparison of our approach with existing unified models.}
        \textbf{Left}: Qualitative comparison illustrating \emph{Mobile-O}'s capabilities in text-to-image generation, visual understanding, and visual prompt understanding. 
        \textbf{Right}: Quantitative comparison with Show-O, Janus, and JanusFlow, demonstrating that \emph{Mobile-O} achieves a superior trade-off. Our \emph{Mobile-O} outperforms Show-O by 5.0\% on GenEval and runs significantly faster on iPhone.
    }
    \label{fig:mobile_o_overview}
\end{center}
}

\maketitle

\maketitle
\renewcommand{\thefootnote}{\fnsymbol{footnote}}
\footnotetext[1]{\scriptsize Corresponding author: \texttt{abdelrahman.youssief@mbzuai.ac.ae}}
\footnotetext[2]{\scriptsize Equal contributions}
\renewcommand{\thefootnote}{\arabic{footnote}}
\clearpage
\begin{abstract}
Unified multimodal models can both understand and generate visual content within a single architecture. 
Existing models, however, remain data-hungry and too heavy for deployment on edge devices. 
We present \emph{Mobile-O}, a compact vision-language-diffusion model that brings unified multimodal intelligence to a mobile device. 
Its core module, the Mobile Conditioning Projector (MCP), fuses vision--language features with a diffusion generator using depthwise-separable convolutions and layerwise alignment. 
This design enables efficient cross-modal conditioning with minimal computational cost. 
Trained on only a few million samples and post-trained in a novel quadruplet format $(\text{generation prompt, image, question, answer})$, \emph{Mobile-O} jointly enhances both visual understanding and generation capabilities. 
Despite its efficiency, \emph{Mobile-O} attains competitive or superior performance compared to other unified models, achieving 74\% on GenEval and outperforming Show-O and JanusFlow by 5\% and 11\%, while running 6$\times$ and 11$\times$ faster, respectively. For visual understanding, \emph{Mobile-O} surpasses them by 15.3\% and 5.1\% averaged across seven benchmarks. 
Running in only $\sim$3s per 512$\times$512 image on an iPhone, \emph{Mobile-O} establishes the first practical framework for real-time unified multimodal understanding and generation on edge devices. We hope \emph{Mobile-O} will ease future research in real-time unified multimodal intelligence running entirely on-device with no cloud dependency. Our code, models, datasets, and mobile application are publicly available.
\end{abstract}    
\section{Introduction}
\label{sec:intro}


Unified multimodal models capable of both \emph{understanding} and \emph{generating} visual content have recently gained popularity in vision. 
Inspired by the success of large language models (LLMs), recent works extend their reasoning and generative capabilities to vision-language tasks, where the unified multimodal models can caption images, answer visual questions, and generate visuals within a single framework~\cite{showo, showo2, blip3o, bagel}. 
Earlier unified approaches~\cite{chameleon,Kim_2023_CVPR} explore a single transformer design that can perform both multimodal understanding and generation, when trained jointly on text and image tokens. 
Subsequent works~\cite{transfusion} incorporate diffusion-based generation directly into unified architectures. Recent methods~\cite{blip3o,bagel}  further explore unified model training on large-scale interleaved multimodal data, achieving improved performance.


Despite these advances, existing unified multimodal models face two critical challenges that limit their practical deployment on consumer devices. 
First, most existing unified models employ computational and memory-demanding visual encoders and denoising modules. For instance, BLIP-3o~\cite{blip3o} requires a $2.6$B-parameter UNet for denoising and $3$B vision-language model (VLM), in addition to $1.5$B for diffusion transformer (DiT), resulting in 7.1B total parameters. While few recent works~\cite{ma2025janusflow} explore computational efficiency in unified multimodal models, they still remain unsuitable for real-time deployment on edge devices (see Fig.~\ref{fig:mobile_o_overview}).  Second, effective cross-modal alignment within unified models often depends on massive pre-training datasets, typically 50M–1B samples~\cite{bagel,blip3o}, making pre-training expensive and time-consuming. 
These observations motivate us to explore a key question: \textit{Can we build a unified multimodal model that is effective for both tasks (understanding and generation), while being efficient for deployment on consumer devices like mobile phones?} 

In this work, we present \textit{Mobile-O}, a compact, efficient unified multimodal model that can run directly on a mobile device with low memory overhead and real-time latency, as shown in Fig.~\ref{fig:mobile_o_overview}. 
Unlike prior approaches that require extensive pre-training, our \textit{Mobile-O} achieves strong understanding and generation performance with only a few million pre-training samples and carefully curated unified post-training data. At the core of our approach is the \textit{Mobile Conditioning Projector}, a mobile-optimized connector that fuses the final hidden states of the VLM with the conditioning space of the diffusion model. 
Furthermore, we address a key limitation in existing training paradigms. Prior unified models either mix disjoint task-specific datasets~\cite{showo,tong2024metamorph} or adopt sequential training that isolates understanding and generation tasks~\cite{blip3o,pan2025transfer}. In contrast, we propose a unified multimodal post-training stage that leverages a compact unified dataset where each sample simultaneously supports \textit{both} tasks through a quadruplet $(\textit{generation prompt, image, question, answer})$ representation for improved cross-modal alignment. 
Finally, we demonstrate real-time deployment of our \textit{Mobile-O} on edge devices, including iPhone, NVIDIA Jetson Nano, and MacBook. The model achieves $\sim$3 seconds per $512\times512$ image generation on an iPhone device, setting a new benchmark for on-device unified multimodal generation. In summary, our key contributions are:

\begin{itemize}
\item We introduce \textit{Mobile-O}, an efficient unified vision–language–diffusion model that achieves state-of-the-art multimodal understanding and image generation performance, while enabling real-time inference on a mobile device (see Fig.~\ref{fig:mobile_o_overview}).
\item To build \textit{Mobile-O}, we first design a solid baseline mobile unified architecture, which is further enhanced with two contributions. First, we introduce the \textit{Mobile Conditioning Projector (MCP)}, a lightweight cross-modal fusion module that effectively bridges visual understanding and diffusion-based generation using depthwise-separable convolutions and layerwise alignment. Second, we propose a unified multimodal post-training scheme that leverages a quadruplet data representation (\textit{generation prompt, image, question, answer}) with a unified dataset of $105k$ samples, enabling joint optimization of multimodal understanding and generation tasks.
\item Our \textit{Mobile-O}, with only 1.6B total parameters, achieves 74\% on \textit{GenEval}, outperforming \textit{Show-O} and \textit{JanusFlow} by 5\% and 11\%, respectively, while being up to 11$\times$ faster. For multimodal image understanding, it surpasses them by 15.3\% and 5.1\%, respectively, on average across seven widely used benchmarks (see Fig.~\ref{fig:mobile_o_overview}).
\end{itemize}
\section{Related Work}
\label{sec:related_work}

\noindent{\textbf{Multimodal Understanding \& Generation:}}

Earlier unified multimodal models ~\cite{lu2024unified, sun2023generative, showo} unify both understanding and generation tasks with a single transformer. Hybrid designs, such as Janus~\cite{janus}, BLIP3-o~\cite{blip3o}, and JanusFlow~\cite{ma2025janusflow} integrate diffusion decoders for better text-to-image generation, while Emu3~\cite{wang2024emu3} shows that auto-regression can suffice for text-to-image generation.

While achieving promising results, the aforementioned unified models either rely on heavy UNet-style~\cite{blip3o} or computationally heavy architectures~\cite{showo,janus} (e.g., CLIP-ViT image encoder).
Moreover, most existing unified models depend on disjoint supervision across understanding and generation, thereby improving one task while freezing the other~\cite{blip3o,pan2025transfer}. 
In contrast, we present a unified mobile-optimized architecture that utilizes a unified multimodal post-training stage where the performance of both tasks is \emph{simultaneously} improved through a multi-task objective.

\noindent{\textbf{Efficient Multimodal Understanding Models:}}
Recent advances in efficient vision-language modeling have focused primarily on optimizing visual encoding strategies~\cite{fastvlm,smolvlm,Shaker2025MobileVideoGPT}. FastVLM~\cite{fastvlm} addresses the computational bottleneck of processing high-resolution images by introducing FastViTHD, a hybrid vision encoder with competitive visual understanding performance. Similarly, SmolVLM~\cite{smolvlm} shows that careful architectural optimizations and aggressive tokenization enable compact models to achieve competitive performance, while consuming less GPU memory. While these approaches focus at efficient multimodal understanding, 
our work advances this research line of efficient multimodal intelligence by introducing a unified framework that couples a compact vision-language understanding model with lightweight diffusion through novel conditioning projector to perform both multimodal understanding and image generation tasks in a single architecture.

\noindent{\textbf{Efficient Text-to-Image Generation Models:}}
Recent works~\cite{sana, snapgen} have explored efficient text-to-image (T2I) generation. SANA~\cite{sana} introduces high-resolution image generation through deep compression autoencoders and linear attention mechanisms. However, they use heavy text encoders (i.e, Gemma-2B~\cite{gemma2_2024}). SnapGen~\cite{snapgen} proposes systematic architecture optimization and cross-architecture distillation, generating images efficiently with multiple steps on resource-constrained devices.
Both approaches are designed for T2I generation and lack multimodal understanding capabilities like FastVLM~\cite{fastvlm}. In contrast, our work strives to design a unified mobile-optimized approach that can effectively perform both multimodal understanding and generation tasks within a single framework. 

\noindent{\textbf{Data Efficiency and Training Stages in Unified Models:}}
Training unified multimodal models typically requires extensive datasets. BAGEL~\cite{xu2025bagel} studies emerging properties in unified multimodal pre-training, revealing fundamental insights about data requirements. Existing unified approaches generally follow two training strategies: \textbf{(i)} \emph{Joint Training}: Methods like Metamorph~\cite{tong2024metamorph} and Show-o~\cite{showo} perform multitask learning by mixing data for image understanding and image generation. While joint training allows the two tasks to potentially benefit from each other~\cite{onecat,mmr1}, its effectiveness strongly depends on the \emph{total data size} and the \emph{ratio between understanding and generation samples}. Current unified training datasets often consist of disjoint subsets for each task~\cite{onecat}, e.g., LLaVA-665K for understanding and BLIP3o-60K for generation, which limits the model's ability to learn fully aligned cross-task understanding. \textbf{(ii)} \emph{Sequential Training}: Other unified works~\cite{janus,pan2025transfer} adopt a two-stage approach: first training the VLM, then freezing the backbone and training only the generation module. For instance, BLIP3-o~\cite{blip3o} uses a pre-trained VLM and freezes it in all stages. This strategy preserves understanding capability, while dedicating to enhance generation performance. However, it does not exploit potential cross-task interactions during training to improve both tasks.

To address these limitations, we introduce a unified post-training stage with $105k$ samples, where
each sample simultaneously supports both understanding and generation. Each training sample is formatted as $(\text{generation prompt, image, question, answer})$, enabling the model to learn aligned understanding and generation capabilities during post-training. This unified format allows us to effectively leverage cross-modal transfer while avoiding the task imbalance and inter-task interference.

\section{Method}
\label{sec:method}

\begin{figure*}
    \centering
    \includegraphics[width=\linewidth]{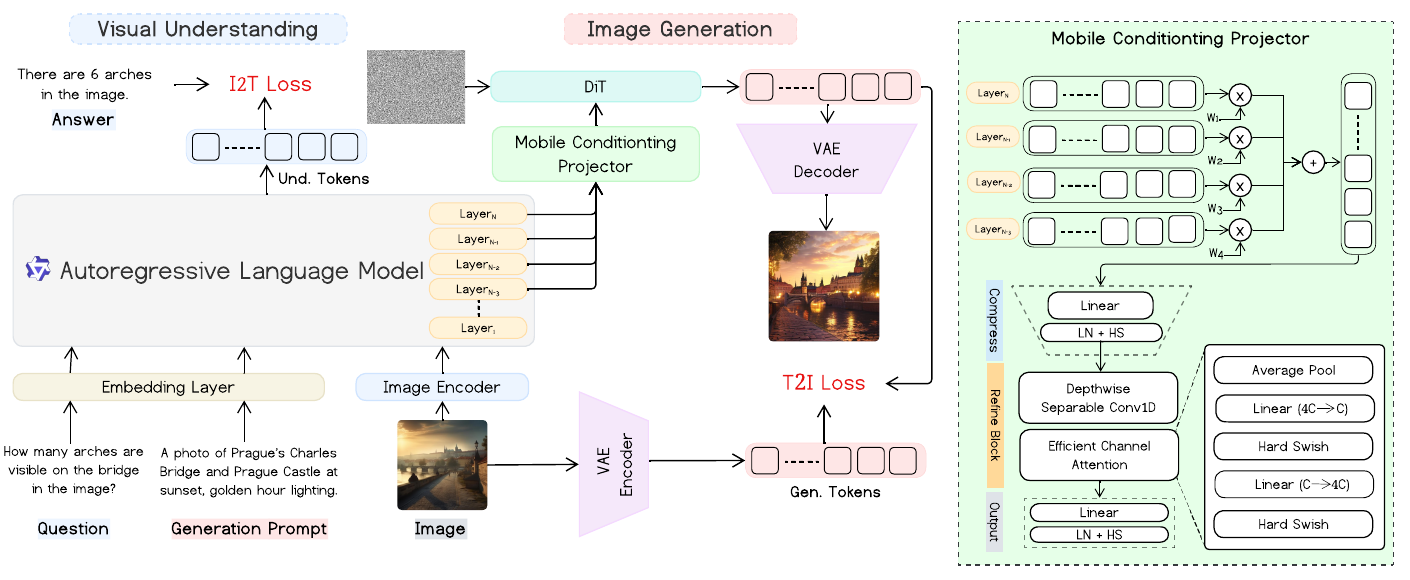}
    \caption{
    \textbf{Overview of \textit{Mobile-O}}. \textbf{Left:} The proposed framework consists of an efficient image encoder with a compact autoregressive language model for visual understanding. For image generation, a lightweight linear diffusion transformer (DiT) is employed alongside a simple yet effective VAE-based encoder–decoder. \textbf{Right:} Our novel Mobile Conditioning Projector (MCP) bridges the understanding and generation tasks by directly conditioning the diffusion model on weighted hidden states from the VLM without the need for intermediate query tokens. The projector leverages layer-wise feature fusion, depthwise separable convolutions, and efficient channel attention to produce high-fidelity conditioning signals with minimal cost, enabling seamless deployment on edge devices.}
  \label{fig:mobile-o-architecture}
    
\end{figure*}

\noindent{\textbf{Motivation:}} To motivate our approach, we distinguish two desirable characteristics to be considered when designing an efficient unified multimodal model for edge deployment.  

\begin{itemize}

\item \textit{\textbf{Efficient Understanding and Generation Connection}}: 
Generally, standard unified models employ a connection module that contains MLP layers to connect understanding and generation components. In addition, the connection module leverages a set of learnable queries that act as a bridge between multimodal LMM and diffusion, enabling improved generation performance. However, such a connection design achieves sub-optimal performance when using substantially less pre-training data (around ~5$\times$ less than BLIP3o~\cite{blip3o}). Therefore, an efficient yet effective connection design is desired to achieve superior performance when constructing a \textit{data-efficient mobile unified} framework.

\item \textit{\textbf{Unified Post-training for Symbiotic Learning}}:
As discussed earlier, most existing unified models either employ joint training~\cite{showo,onecat} or utilize sequential training~\cite{blip3o,janus} for understanding and generation. However, joint training typically relies on a careful balancing of \textit{disjoint} understanding and generation data samples, whereas sequential training only aims to improve \textit{one} task (e.g., generation) while freezing the other (e.g., understanding). To address this, a unified post-training approach is desired based on a multi-task objective using a \textit{joint} set of understanding and generation data samples to simultaneously improve \textit{both} understanding and generation tasks. 

\end{itemize}

\subsection{Baseline Mobile Unified Framework}
Since existing mobile-optimized models are designed to either perform multimodal visual understanding or image generation, we first aim at building a solid baseline mobile unified architecture capable of handling both tasks. Motivated by recent unified models such as BLIP-3o~\cite{blip3o}, which build generation capabilities directly on top of existing understanding models (e.g., Qwen2-VL), we adopt a similar yet mobile-optimized design strategy. To establish a strong mobile unified baseline, we consider efficient pre-trained vision-language model (VLM) backbones and diffusion decoders in configurations reflecting prior unified models. Specifically, as our baseline, we employ FastVLM~\cite{fastvlm} for multimodal understanding and integrate it with a DiT-style diffusion decoder~\cite{sana} for multimodal generation.

Let $f_{\theta}$ denote the vision-language encoder-decoder (FastVLM~\cite{fastvlm}) and $g_{\phi}$ the diffusion image decoder (SANA-0.6B~\cite{sana}).  
Given a text prompt $p$ and an optional image $x$ (for understanding), the VLM produces layerwise hidden states 
$\{H^{(1)},\dots,H^{(L)}\}$, where $H^{(\ell)}\in\mathbb{R}^{N\times d_{\text{vlm}}}$ for token length $N$ and hidden size $d_{\text{vlm}}$.  The diffusion model $g_{\phi}$ is a DiT-style decoder with cross-attention blocks accepting encoder features of dimension $d_{\text{cond}}$.  
Following recent unified models~\cite{showo,showo2,blip3o,tbac}, $g_{\phi}$ remains fully learnable, but we avoid introducing extra textual tokens beyond those produced by $f_{\theta}$. Unlike SANA-0.6B~\cite{sana}, which uses the Gemma-2B~\cite{gemma2_2024} model as a text encoder to process generation prompts, we employ the same LLM used for the understanding model to handle the generation prompts, resulting in a more parameter-efficient design.

Our goal is to jointly learn $\theta$ and $\phi$ so the model can (i) perform visual understanding tasks (e.g., question answering) and (ii) generate images from prompts, all within a mobile-optimized architecture. Next, we discuss how to further improve the performance of the baseline mobile unified framework through an efficient yet effective projector design and a unified post-training approach with a multi-task objective to improve understanding and generation. 

\subsection{Mobile Conditioning Projector (MCP)}
\label{sec:mcp}
Unified frameworks usually insert learnable query tokens between the VLM and the image decoder~\cite{blip3o,tbac,pan2025transfer}.
While this approach is effective for large models, it requires massive pre-training data for effective alignment. To this end, we design an efficient yet effective conditioning projection (MCP) layer that directly connects VLM hidden states to the diffusion decoder, as shown in Fig.~\ref{fig:mobile-o-architecture}.
The MCP maps the VLM’s final-layer features (or a fusion of the last $K$ layers) to diffusion-compatible conditioning sequences with \emph{minimal} parameters and FLOPs.

\noindent{\textbf{Layerwise Fusion.}}
Let $\mathcal{S}=\{L{-}K{+}1,\dots,L\}$ denote the last $K$ VLM layers.
We compute a temperature-scaled softmax weighting $\alpha_{\ell}=\frac{\exp(w_{\ell}/\tau)}{\sum_{j\in\mathcal{S}}\exp(w_{j}/\tau)}$, and form a fused representation,
\begin{equation}
\label{eq:fuse}
H_{\text{fuse}} \;=\; \sum_{\ell\in\mathcal{S}} \alpha_{\ell}\, H^{(\ell)} 
\;\in\; \mathbb{R}^{N\times d_{\text{vlm}}}\,.
\end{equation}
where the weights $\{w_{\ell}\}$ are learned; $\tau$ is cosine-annealed during the training.

\noindent{\textbf{Compression and Refinement.}}
We project $H_{\text{fuse}}$ to a compact space and refine it using \emph{depthwise-separable 1D convolutions} and \emph{lightweight channel attention}:
\begin{align}
\label{eq:compress}
\tilde{H} &= \mathrm{LN}\!\big(H_{\text{fuse}} W_c\big), \qquad W_c\in\mathbb{R}^{d_{\text{vlm}}\times d_h},\\
\label{eq:refine}
\tilde{H} &\leftarrow \mathrm{SeqRefine}\big(\tilde{H}\big),
\end{align}
where $\mathrm{SeqRefine}$ applies a depthwise-separable $\text{Conv1D}$ followed by pointwise mixing and a tiny MLP-based channel attention.
Operating along sequence length $N$ (not spatial grids) avoids expensive 2D convolutions and retains token-level alignment with language stream.

\noindent\textbf{Output Projection.}
The diffusion cross-attention expects $d_{\text{cond}}$-dimensional keys and values. We compute
\begin{equation}
\label{eq:proj}
E = \mathrm{LN}(\tilde{H} W_o), \quad 
W_o\!\in\!\mathbb{R}^{d_h\times d_{\text{cond}}}, \;
E\!\in\!\mathbb{R}^{N\times d_{\text{cond}}}.
\end{equation}

All cross-attention layers in $g_{\phi}$ use the \emph{same} sequence $E$ as encoder features, analogous to CLIP-conditioning in latent diffusion, but learned \emph{end-to-end} with the VLM. 
Compared to query-token approaches~\cite{blip3o,tbac,pan2025transfer}, the proposed MCP introduces no extra token budget and reduces parameter count and requires less pre-training data.

\noindent{\textbf{Complexity.}}
For hidden size $d_h$ and kernel $k$, the refinement block costs $\mathcal{O}(k\,d_h)$ (depthwise) $+\ \mathcal{O}(d_h^2)$ (pointwise) per token, substantially cheaper than full 2D convolution or attention over new query tokens. 

\begin{figure}[t]
    \centering
    \includegraphics[width=\linewidth]{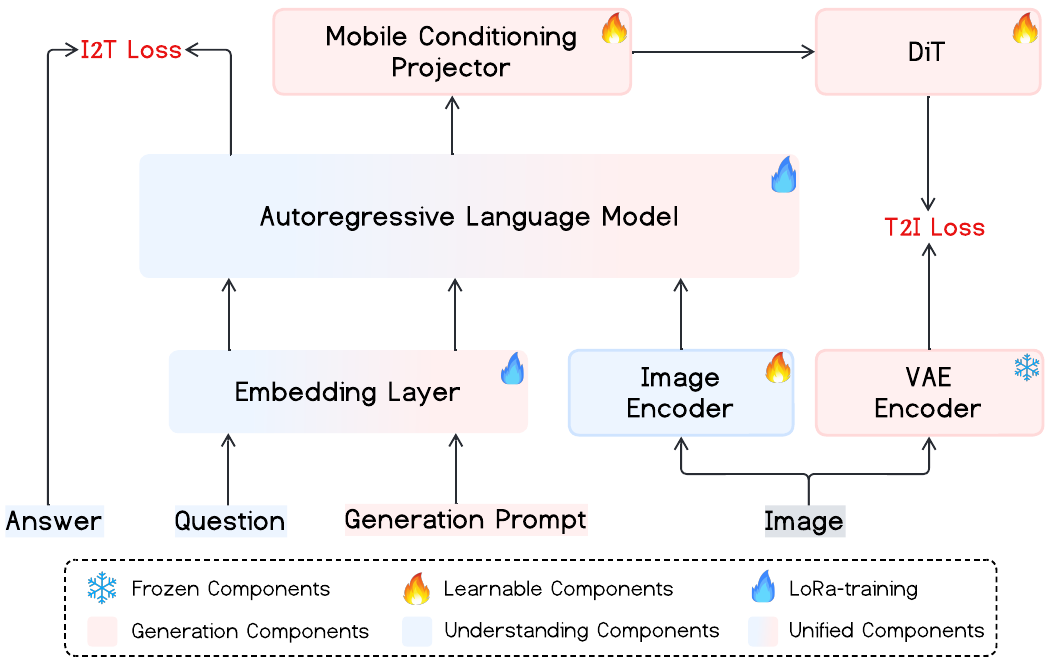}
    \caption{\textbf{Overview of the proposed unified multimodal post-training pipeline.} We jointly optimize multimodal understanding and generation through a multi-task objective using a quadruplet format (\textit{generation prompt, image, question, answer}). Both I2T and T2I losses are computed simultaneously, enabling aligned cross-modal learning where each training sample supports both multimodal understanding and generation.}
    \label{fig:mobile-o-architecture_UnifiedTr}
\end{figure}

\begin{table*}[t]
    \centering
    \caption{\textbf{Comparison with recent multimodal understanding models.} ``Und.'' and ``Gen.'' denote ``understanding'' and ``generation'', respectively. \textbf{Total Params} represent the sum of visual encoder, language model, and diffusion/unet components (when applicable). Compared to unified models with similar size ($\leq$ 2B), our \emph{Mobile-O-0.5B} achieves superior overall performance with a score of 61.9 averaged over seven datasets. Further, \emph{Mobile-O-0.5B} also outperforms its understanding-only counterpart (FastVLM) by 1.6\% in average performance. }
    \resizebox{\textwidth}{!}{%
    \begin{tabular}{llcccccccccc}
        \toprule
        \textbf{Type} & \textbf{Model} & \textbf{\# Total Params} & \textbf{MMMU$ \uparrow$} & \textbf{TextVQA$ \uparrow$} & \textbf{MMVet$ \uparrow$} & \textbf{SEED$ \uparrow$} & \textbf{ChartQA$ \uparrow$} & \textbf{POPE$ \uparrow$} & \textbf{GQA$ \uparrow$} & \textbf{Average$ \uparrow$} \\
        \midrule
        \textit{Und. Only $>$ 1B} & LLaVA-Phi~\cite{zhu2024llava} & $3.1$B & - & $48.6$ & $28.9$ & - & - & $85.0$ & - & - \\
        & LLaVA-v$1.5$-Phi-$1.5$~\cite{zhu2024llava} & $1.6$B & $30.7$ & - & - & - & - & $84.1$ & $56.5$ & - \\
        & MobileVLM~\cite{mobilevlm_v1} & $1.7$B & - & $41.5$ & - & - & - & $84.5$ & $56.1$ & - \\
        & MobileVLM-V2~\cite{mobilevlm_v2} & $1.7$B & - & $52.1$ & - & - & - & $84.3$ & $59.3$ & - \\
        & LLaVa-OV~\cite{li2025llavaonevision} & $1.6$B & $31.4$ & - & $29.1$ & $65.5$ & $61.4$ & - & - & - \\ \dashedmidrule
        \textit{Und. Only $\leq$ 1B} & Smol-VLM-0.5B~\cite{smolvlm} & $0.6$B & $33.7$ & $60.2$ & - & - & $62.8$ & - & - & - \\
        & FastVLM-0.5B~\cite{fastvlm} & $0.6$B & $33.3$ & $68.0$ & $37.5$ & $69.3$ & $71.6$ & $81.1$ & $62.7$ & $60.5$ \\
        \midrule
        \textit{Und. and Gen. $>$ 2B} & EMU3-8B~\cite{wang2024emu3} & $9.0$B & $31.6$ & $64.7$ & $37.2$ & $68.2$ & $68.6$ & $85.2$ & $60.3$ & $59.4$ \\
        & BLIP3o-4B~\cite{blip3o} & $7.1$B & $46.6$ & $78.0$ & $60.1$ & $73.8$ & - & - & - & - \\ \dashedmidrule
        \textit{Und. and Gen. $\leq$ 2B} & Janus~\cite{janus} & $2.1$B & $30.5$ & $50.2$ & $34.3$ & $63.7$ & $53.0$ & $87.0$ & $59.1$ & $54.0$ \\
        & Show-o~\cite{showo} & $1.5$B & $25.1$ & - & - & - & - & $73.8$ & $48.7$ & - \\
        & Show-o-Clip-ViT~\cite{showo} & $1.6$B & $27.4$ & $41.2$ & $20.9$ & $51.6$ & $44.7$ & $84.5$ & $57.5$ & $46.8$ \\
        & JanusFlow~\cite{ma2025janusflow} & $2.1$B & $29.3$ & $55.5$ & $30.9$ & $70.5$ & $64.6$ & $88.0$ & $60.3$ & $57.0$ \\ \rowcolor{front-color}
        & \textbf{Mobile-O-0.5B (Ours)} & $\mathbf{1.6}$B & $34.6$ & $67.8$ & $38.1$ & $69.4$ & $75.2$ & $86.4$ & $62.9$ & $\mathbf{62.1}$ \\
        \bottomrule
    \end{tabular}
    }
    \label{sota_result_understanding_unified}
\end{table*}

\subsection{Training Scheme}
\label{sec:training_strategy}

We propose a three-stage training scheme for our \textit{Mobile-O} that progressively enhances multimodal understanding and generation capabilities. The three stages are: cross-modal alignment, supervised fine-tuning and unified multimodal post-training. During the first two stages, the visual encoders and LLM backbone are frozen to learn better multimodal generation. The focus of our design is the introduction of a novel unified multimodal post-training stage (stage 3), where both multimodal understanding and generation are improved using a joint set of data samples via a multi-task objective (see Fig.~\ref{fig:mobile-o-architecture_UnifiedTr}).  

\noindent{\textbf{Stage 1: Cross-Modal Alignment}}. Here, the primary objective is to establish robust connections between visual and linguistic representations within a unified embedding space. We adopt a parameter-efficient approach by freezing the visual encoders and LLM backbone, and update only the DiT and MCP. In this stage, we conduct pre-training on JourneyDB~\cite{sun2023journeydb}, which provides high-quality 4 million text–image pairs covering diverse visual concepts, and 5 million pairs from BLIP3o-Short-Caption~\cite{blip3o}, a curated subset emphasizing compositional understanding.

\noindent{\textbf{Stage 2: Supervised Fine-tuning}}. Following initial alignment, we perform targeted fine-tuning on $\sim$105K curated prompt-image pairs (60K from BLIP3o~\cite{blip3o}, 45K from ShareGPT-4o-Image~\cite{chen2025sharegpt4oimage}) to address specific weaknesses observed after pre-training~\cite{blip3o}. Due to our compact pre-training corpus (only 20\% of BLIP-3o's data during stage 1), the model initially struggled with complex human gestures, common objects and landmarks. This stage specifically targets these underrepresented domains while maintaining the same frozen/trainable component configuration as in the previous stage.

\noindent{\textbf{Stage 3: Unified Multimodal Post-Training}}.
This stage aims to improve both multimodal understanding and generation. To this end, we
construct training samples as quadruplets $\mathcal{S} = \{p, \mathbf{x}_{\text{img}}, q, a\}$, where $p$ denotes the generation prompt, $\mathbf{x}_{\text{img}}$ represents the image, and $(q, a)$ form question-answer pairs (see Fig.~\ref{fig:mobile-o-architecture_UnifiedTr}). Since no existing dataset supports such a quadruplet format, we construct the data as  follows:
\begin{enumerate}[noitemsep,topsep=0pt]
    \item Prompt GPT-4o~\cite{openai2024gpt4o} to generate highly detailed compositionally-aware caption for each image.
    \item Synthesize diverse question-answer sets probing different aspects of visual understanding.
\end{enumerate}

This yields a unified dataset with bi-directional multimodal learning within a single framework, where both understanding with image-to-text (I2T) and generation with text-to-image (T2I) tasks share the same embedding layer and autoregressive language model, as shown in Fig.~\ref{fig:mobile-o-architecture_UnifiedTr}.

\subsection{Training Objectives}
\label{sec:loss}

Our unified training optimizes a weighted combination of multimodal understanding and generation objectives:

\begin{equation}
\label{eq:total}
\mathcal{L}_{\text{unified}} = \lambda_{\text{lang}} \mathcal{L}_{\text{lang}} + \lambda_{\text{diff}} \mathcal{L}_{\text{diff}}
\end{equation}

\noindent{\textbf{Image-to-Text (I2T) Loss}.}
For multimodal understanding, we employ standard cross-entropy loss on the autoregressive language model's output tokens:
\begin{equation}
\mathcal{L}_{\text{lang}} = -\sum_{t=1}^{|a|} \log P(a_t | \mathbf{x}_{\text{img}}, q, a_{<t})
\end{equation}
where, the model predicts answer tokens $a$ conditioned on the image encoding and question $q$.

\noindent{\textbf{Text-to-Image (T2I) Loss}.} For multimodal image generation, we employ a flow-matching objective~\cite{showo2,blip3o} instead of standard noise prediction. Given a clean latent $\mathbf{x}$ from the VAE encoder and noise $\epsilon \sim \mathcal{N}(0,I)$, we sample a noise level $\sigma \in [0,1]$ and form:
\begin{equation}
\mathbf{x}_{\sigma} = (1-\sigma)\mathbf{x} + \sigma\epsilon, \quad v^{\star}(\mathbf{x}_{\sigma};\sigma) = \epsilon - \mathbf{x}
\end{equation}

The DiT model predicts a velocity field $v_{\phi}(\mathbf{x}_{\sigma}, \sigma, \mathbf{c}_p)$ conditioned on MCP features $\mathbf{c}_p$ derived from the generation prompt $p$ (see Eq.~\ref{eq:proj}). The loss minimizes the weighted mean-squared error:
\begin{equation}
\label{eq:flowmatch}
\mathcal{L}_{\text{diff}} = \mathbb{E}_{\mathbf{x},p,\epsilon,\sigma} \left[ w(\sigma) \left\| v_{\phi}(\mathbf{x}_{\sigma}, \sigma, \mathbf{c}_p) - (\epsilon - \mathbf{x}) \right\|_2^2 \right]
\end{equation}
where, $w(\sigma)$ is a scale-dependent weighting function. This formulation directly learns the probability flow ODE, yielding faster and more stable training compared to standard diffusion objectives.


\section{Experiments}
\label{sec:results}

\subsection{Implementation Details}
We use FastVLM-0.5B~\cite{fastvlm} as the image understanding model, which extends FastViT~\cite{vasu2023fastvit} as the vision encoder and Qwen2-0.5B~\cite{qwen2_technical_report} as the language backbone. For image generation, we adopt the SANA-600M-512~\cite{sana} diffusion model as the visual generator. Both understanding and generation branches are connected through the proposed Mobile Conditioning Projector, implemented as a lightweight linear layers with depthwise-separable convolutions for efficient cross-modal alignment. All images used for understanding tasks are resized to 1024 $\times$ 1024 resolution using bicubic interpolation, while generation tasks operate at 512 $\times$ 512. All experiments are conducted on a single node equipped with 8 NVIDIA A100 GPUs, requiring approximately 3 days for 50k pre-training steps (roughly 3 epochs). The subsequent SFT and unified multimodal post-training stages run for 20 epochs and 7 epochs, taking 15 hours and 5 hours, respectively. Detailed hyperparameter configurations for each stage are provided in the suppl. material.

\begin{table*}[t]
    \centering
    \caption{\textbf{Evaluation of text-to-image generation performance on the GenEval benchmark.} ``Und.'' and ``Gen.'' denote ``understanding'' and ``generation'', respectively. \textbf{ Total Params} represent the sum of the visual encoder, language model, and diffusion/unet components (when applicable). Compared to unified models with similar size ($\leq$ 2B), our \emph{Mobile-O-0.5B} achieves superior overall score of 0.74 and outperforms Show-o-Clip-ViT~\cite{showo} by 5.0\%.}
    \label{tab:geneval_unified}
    \resizebox{\textwidth}{!}{%
    \begin{tabular}{llcccccccc}
        \toprule
        \textbf{Type} & \textbf{Method} & \textbf{\# Total Params} & \textbf{Single Obj.} & \textbf{Two Obj.} & \textbf{Counting} & \textbf{Colors} & \textbf{Position} & \textbf{Color Attri.} & \textbf{Overall$\uparrow$} \\
        \midrule
        \textit{Gen. Only $>$ 1B} & LlamaGen~\cite{sun2024autoregressive} & $3.8$B & $0.71$ & $0.34$ & $0.21$ & $0.58$ & $0.07$ & $0.04$ & $0.32$ \\
        & LDM~\cite{rombach2022high} & $1.5$B & $0.92$ & $0.29$ & $0.23$ & $0.70$ & $0.02$ & $0.05$ & $0.37$ \\
        & PixArt-$\alpha$~\cite{chen2023pixart} & $4.9$B & $0.98$ & $0.50$ & $0.44$ & $0.80$ & $0.08$ & $0.07$ & $0.48$ \\
        & SDXL~\cite{podell2024sdxl} & $3.0$B & $0.98$ & $0.74$ & $0.39$ & $0.85$ & $0.15$ & $0.23$ & $0.55$ \\
        & SDv$2.1$~\cite{rombach2022high} & $1.9$B & $0.98$ & $0.51$ & $0.44$ & $0.85$ & $0.07$ & $0.17$ & $0.50$ \\
        & SANA-0.6B~\cite{sana} & $2.6$B & $0.99$ & $0.77$ & $0.62$ & $0.88$ & $0.21$ & $0.47$ & $0.66$ \\ \dashedmidrule
        \textit{Gen. Only $\leq$ 1B} & SDv$1.5$~\cite{rombach2022high} & $1.0$B & $0.97$ & $0.38$ & $0.35$ & $0.76$ & $0.04$ & $0.06$ & $0.43$ \\
        & SnapGen~\cite{snapgen} & $0.4$B & $1.00$ & $0.84$ & $0.60$ & $0.88$ & $0.18$ & $0.45$ & $0.66$ \\ \midrule
        \textit{Und. and Gen. $>$ 2B} & SEED-X$^\dagger$~\cite{ge2024seed} & $16.0$B & $0.97$ & $0.58$ & $0.26$ & $0.80$ & $0.19$ & $0.14$ & $0.49$ \\
        & Chameleon~\cite{chameleon} & $34.0$B & - & - & - & - & - & - & $0.39$ \\
        & LWM~\cite{liu2024world} & $7.2$B & $0.93$ & $0.41$ & $0.46$ & $0.79$ & $0.09$ & $0.15$ & $0.47$ \\
        & BLIP3o-4B~\cite{blip3o} & $7.3$B & - & - & - & - & - & - & $0.81$ \\ \dashedmidrule
        \textit{Und. and Gen. $\leq$ 2B} & Janus~\cite{janus} & $2.1$B & $0.97$ & $0.68$ & $0.30$ & $0.84$ & $0.46$ & $0.42$ & $0.61$ \\
        & Show-o~\cite{showo} & $1.5$B & $0.98$ & $0.80$ & $0.66$ & $0.84$ & $0.31$ & $0.50$ & $0.68$ \\
        & Show-o-Clip-ViT~\cite{showo} & $1.6$B & $0.98$ & $0.85$ & $0.67$ & $0.81$ & $0.28$ & $0.55$ & $0.69$ \\
        & JanusFlow~\cite{ma2025janusflow} & $2.1$B & $0.97$ & $0.59$ & $0.45$ & $0.83$ & $0.53$ & $0.42$ & $0.63$ \\ \rowcolor{front-color}
        & \textbf{Mobile-O-0.5B (Ours)} & $\mathbf{1.6}$B & $0.98$ & $0.87$ & $0.57$ & $0.86$ & $0.68$ & $0.49$ & $0.74$ \\
        \bottomrule
    \end{tabular}
    }
\end{table*}

\begin{figure*}
    \centering
    \includegraphics[width=\linewidth]{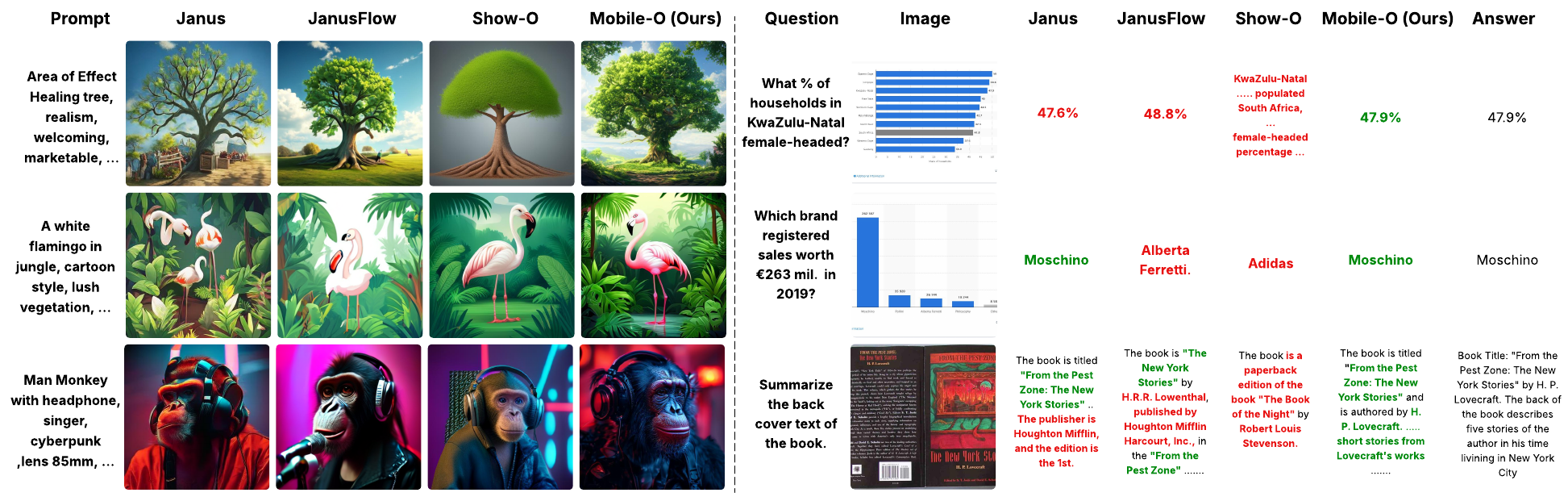}
    \caption{\textbf{Qualitative comparison of text-to-image generation (left) and visual understanding (right) across unified multimodal models}. Each column shows Janus, JanusFlow, Show-O, and Mobile-O (ours) for the same prompts/questions. Mobile-O yields more consistent, detailed, and semantically faithful images with high fidelity and style diversity for image generation. For visual understanding, it delivers more accurate and contextually coherent responses. Additional results are presented in suppl. material. Best viewed zoomed in.}
    \label{fig:qualitative_comparison}
\end{figure*}

\subsection{Quantitative Comparison}

\noindent{\textbf{Multimodal Visual Understanding:}}
We evaluate \emph{Mobile-O-0.5B} on a diverse suite of understanding benchmarks. General multimodal understanding and reasoning are evaluated on MMMU~\cite{yue2024mmmu}, MM-Vet~\cite{yu2023mmvet}, and SEED~\cite{li2023seedbench}. For OCR and text-based VQA, we employ TextVQA~\cite{singh2019textvqa} and ChartQA~\cite{masry2022chartqa}. Text hallucination robustness is examined on POPE~\cite{li2023pope}, while scene understanding is assessed on GQA~\cite{hudson2019gqa}. 
Tab.~\ref{sota_result_understanding_unified} shows the comparison with understanding-only models having $<$1B and $>$1B and unified  models with $<$2B and $>$2B, on seven benchmarks.
Here, the total number of parameters reflects all components, and not only the LLM. \textit{Mobile-O-0.5B} offers distinct merits over models in its scale range ($\leq$2B), such as Janus~\cite{janus}, JanusFlow~\cite{ma2025janusflow}, and Show-O~\cite{showo}. Compared to JanusFlow~\cite{ma2025janusflow}, our \emph{Mobile-O-0.5B} obtains an absolute gain of 4.9\% averaged over seven benchmarks with less total parameters (JanusFlow: 2.1B vs. Ours: 1.6B). It is worth mentioning that our \emph{Mobile-O-0.5B} obtains an absolute gain of 1.6\% over FastVLM~\cite{fastvlm}, highlighting the effectiveness of our unified multimodal post-training, where both understanding and generation tasks are improved via a multi-task objective using joint training samples as quadruplets.

\noindent{\textbf{Text-to-Image Generation:}}
We evaluate our model on the widely-used GenEval~\cite{ghosh2023geneval} benchmark. We follow strictly to raw prompts for GenEval. 
As shown in Tab.~\ref{tab:geneval_unified}, we evaluate \emph{Mobile-O-0.5B} with generation-only models having different sizes ($>$ 1B and $\leq$ 1B) and unified  models ( $>$ 2B and $\leq$ 2B). Here, total number of parameters reflects all components.
Compared to unified models with similar size ($\leq$ 2B), our \emph{Mobile-O-0.5B} achieves best overall results with score of 0.74, outperforming Show-o~\cite{showo} by 5.0\%. 

\noindent{\textbf{Text-and-Image-to-Image Generation}: Beyond text-to-image generation and visual understanding, the Mobile-O framework naturally supports \textit{image editing}, taking both a source image and a textual instruction as input and producing an edited image as output. This capability emerges from the MCP design, which bridges the understanding and generation pathways through a shared multimodal representation. Because MCP captures low-level visual details from the input image, it is well-suited for editing tasks that require preserving the global scene structure while applying localized modifications.

To enable image editing, we fine-tune Mobile-O on a small subset of 46k editing samples from ShareGPT4V~\cite{chen2025sharegpt4oimage}. During editing, the source image is encoded through the vision encoder and projected via MCP, while the textual editing instruction is processed by the language model. The generation backbone then produces the edited image conditioned on both the visual and textual representations. No architectural modifications are required—the same MCP, language model, and generation backbone used for text-to-image generation and visual understanding are reused for editing. We evaluate \emph{Mobile-O-0.5B} on the ImageEdit~\cite{ye2025imgedit} benchmark, which measures both edit fidelity and scene preservation. \emph{Mobile-O-0.5B} achieves an overall score of {2.5} on ImageEdit, despite being fine-tuned on only 46k editing samples.  We note that \emph{Mobile-O-0.5B's} editing capability is achieved with minimal dedicated training data compared to specialized editing models such as BLIP3-o~\cite{blip3o} and Emu-Edit~\cite{wang2024emu3}, which are trained on significantly larger editing-specific datasets. With dedicated fine-tuning on larger-scale editing data, we expect both the edit fidelity and global scene preservation to further improve. 

\begin{figure}[t!]
\centering
\includegraphics[width=\columnwidth]{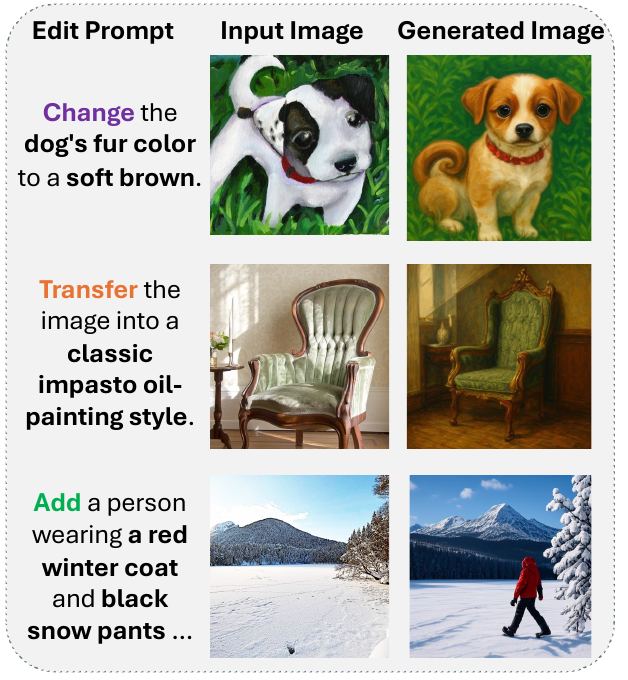}
\caption{Qualitative image editing results of Mobile-O-0.5B. Given a source image and a textual editing instruction, Mobile-O-0.5B produces the edited output. The model is fine-tuned on only 46k editing samples from ShareGPT4V~\cite{chen2025sharegpt4oimage}.}
\label{fig:image_editing_supp}
\end{figure}

\subsection{Qualitative Comparison}

Fig.~\ref{fig:qualitative_comparison} illustrates the generation and understanding capabilities of \textit{Mobile-O-0.5B} with other unified models $\leq$ 2B parameters. Compared to Janus, JanusFlow, and Show-O, \textit{Mobile-O-0.5B} produces images with sharper details, more coherent layouts, and more consistent illumination. It maintains higher visual fidelity in complex scenes, such as tree leaves or strands of a monkey’s hair. Janus and JanusFlow show counting errors in the second row of Fig.~\ref{fig:qualitative_comparison}, consistent with their lower counting scores in Tab.~\ref{tab:geneval_unified}. These counting issues sometimes yield higher diversity but reduce text–image alignment. For understanding, \textit{Mobile-O-0.5B} correctly answers samples from ChartQA~\cite{masry2022chartqa} and TextVQA~\cite{singh2019textvqa}, and in the last row accurately summarizes a book cover, mentioning both title and author.
Complete output comparison is provided in suppl. material. In Fig.~\ref{fig:image_editing_supp}, Mobile-O-0.5B successfully performs a range of basic editing operations, including adding an object, attribute modification, and style transfer.

\subsection{Ablation Study}
\label{sec:ablation}

\noindent\textbf{Generality of Mobile-O:} A natural question is whether the Mobile-O framework, specifically the Multi-modal Connector Projector (MCP), unified post-training data format, and training recipe, generalizes beyond the specific backbone choices presented in the main paper. To address this, we construct {Mobile-O-1.5B} by replacing the original components with larger counterparts: FastVLM-1.5B~\cite{fastvlm} as the vision-language understanding backbone and SANA-1.5B~\cite{sana} as the image generation backbone, yielding a unified model with approximately 3.5B parameters. The MCP dimensions are adjusted accordingly to match the hidden sizes of the larger backbones, while the overall architecture and training procedure remain unchanged. We evaluate understanding performance across seven established benchmarks: MMMU~\cite{yue2024mmmu}, TextVQA~\cite{singh2019textvqa}, SEED-Bench~\cite{li2023seedbench}, ChartQA~\cite{masry2022chartqa}, POPE~\cite{li2023pope}, GQA~\cite{hudson2019gqa}, and MM-Vet~\cite{yu2023mmvet}. For generation quality, we report the GenEval~\cite{ghosh2023geneval} overall score. Results are summarized in Table~\ref{tab:results_rebuttal_supp}.

Mobile-O-1.5B after supervised fine-tuning preserves the full understanding capability of the standalone FastVLM-1.5B (64.8\% average across the seven benchmarks) while simultaneously gaining strong generation ability (75\% GenEval), which the original FastVLM entirely lacks. After the post-training stage, both capabilities improve further: understanding increases to {66.2\%} (+1.4\% absolute over SFT) and generation reaches {78\%} (+3\% absolute over SFT). Notably, the post-trained Mobile-O-3B also surpasses the standalone SANA-1.5B generation backbone (78\% vs.\ 66\%), demonstrating that the unified training and post-training recipe not only preserves but enhances the individual component capabilities. These results confirm that the Mobile-O framework is \textit{architecture-agnostic}: the MCP design, unified data format, and post-training recipe transfer effectively to larger backbones, consistently improving both understanding and generation. 

\begin{table}[t!]
\centering
\caption{\textbf{Mobile-O-1.5B}: Scaling to FastVLM-1.5B and SANA-1.5B components. Understanding performance is averaged over seven benchmarks (MMMU, TextVQA, SEED-Bench, ChartQA, POPE, GQA, MM-Vet). Generation quality is measured by GenEval overall score. The proposed post-training stage consistently improves both capabilities.}

\resizebox{\columnwidth}{!}{%
\begin{tabular}{lcc}
\toprule
\textbf{Model} & \textbf{Und. Acc. (\%)} & \textbf{Gen. Acc. (\%)} \\
\midrule
FastVLM-1.5B~\cite{fastvlm} & 64.8\% & -- \\
SANA-1.5B~\cite{sana} & -- & 66\% \\
Mobile-O-1.5B (SFT) & 64.8\% & 75\% \\
\rowcolor{front-color}
\textbf{Mobile-O-1.5B (Post-train)} & \textbf{66.2\%} & \textbf{78\%} \\
\bottomrule
\end{tabular}%
}

\label{tab:results_rebuttal_supp}
\end{table}

We analyze the contributions of the proposed MCP design and the effectiveness of our post-training data strategy.

\noindent\textbf{On the MCP Design.}
Tab.~\ref{tab:mcp_ablation} shows how different MCP configurations influence cross-modal alignment and generation quality. Notably, all experiments in this table are conducted without pre-training. Using a simple MLP connector between the VLM and diffusion decoder achieves 68.5\% on GenEval but requires over 3.2M trainable parameters. Replacing it with our single-layer MCP with a compression module reduces parameter count by nearly half, while maintaining comparable performance of 68.4\%. Extending to the last four layers with uniform fusion further improves alignment to 69.6\%. Introducing learnable weights across layers enables the model to dynamically attend to informative representations, boosting accuracy to 70.0\%. Finally, adding the lightweight refinement block leads to best results of 70.4\% with only 2.4M parameters.

{
\setlength{\textfloatsep}{1pt} 
\setlength{\intextsep}{1pt}    
\begin{table}[t]
\centering
\caption{
\textbf{Ablation on the Mobile Conditioning Projector (MCP).}
We study the effect of layer fusion, learnable weighting, and the refinement block.}
\label{tab:mcp_ablation}
\resizebox{\columnwidth}{!}{
\begin{tabular}{lcccccc}
\toprule
\textbf{Proj.} & \textbf{\# Layers} & \textbf{Fusion} & \textbf{Compress} & \textbf{CA} & \textbf{Acc. (\%)} & \textbf{Params (M)} \\
\midrule
MLP  & -- & -- & -- & -- & 68.5 & 3.3 \\
MCP  & 1 & Uniform & \checkmark & -- & 68.4 & 1.7 \\
MCP  & 4 & Uniform & \checkmark & -- & 69.6 & 1.7 \\
MCP  & 4 & Learnable & \checkmark & -- & 70.0 & 1.7 \\
\rowcolor{front-color}
MCP & 4 & Learnable & \checkmark & \checkmark & 70.4 & 2.4 \\
\bottomrule
\end{tabular}
}
\end{table}
}

\noindent\textbf{On the Effect of Unified Post-Training.}
Tab.~\ref{tab:posttrain_ablation} evaluates our efficient post-training phase designed to enhance both understanding and generation tasks. We compare standard SFT against two post-training variants. Adding post-training with generation-only triplets slightly improves results across benchmarks, showing better consistency in generative alignment. When generation and understanding triplets are used jointly, we observe measurable improvements, increasing average accuracy on seven image understanding tasks from 60.5\% to 62.1\% and GenEval by 1\%. These results demonstrate that multi-objective post-training is a straightforward yet effective approach to enhance cross-modal coherence without need for large-scale pre-training.

\begin{table}[t]
\centering
\caption{
\textbf{Effect of Unified Post-Training.}
Our post-training data improves both understanding and generation alignment when using joint quadruplets.}
\label{tab:posttrain_ablation}
\resizebox{\columnwidth}{!}{
\begin{tabular}{lccc}
\toprule
\textbf{Method} & \textbf{Und. Acc. (\%)} & \textbf{Gen. Acc. (\%)}  \\
\midrule
SFT & 60.5 & 73.3 \\
SFT + Post-Train (image-text pairs) & 60.6 & 73.4 \\
\rowcolor{front-color}
SFT + Post-Train (quadruplets) & \textbf{62.1} & \textbf{74.2}  \\
\bottomrule
\end{tabular}
}
\end{table}

\subsection{Edge Deployment}
To assess the practicality on consumer devices, we evaluate recent unified methods below 2B parameters on three representative edge platforms: MacBook M2 Pro, NVIDIA Jetson Orin Nano, and iPhone 17 Pro. Tab.~\ref{tab:image_understanding_generation} reports inference times for \textit{visual understanding} (vision encoder + text token forward time, TTFT) and total latency for \textit{image generation} with 20 denoising steps. \emph{Mobile-O-0.5B} demonstrates notable efficiency gains over prior unified models. On the MacBook M2 Pro, it is $2$–$8\times$ faster than Janus and Show-O for understanding and $11$–$46\times$ faster for image generation. On Jetson Orin Nano, \emph{Mobile-O-0.5B} generates images in only 4 s, vs. 22–52 s for other methods. On iPhone 17 Pro, \emph{Mobile-O-0.5B} achieves vision encoder latency of 102 ms, TTFT of 248 ms, and image generation in 3.0 s, highlighting its suitability for real-world deployment.

For mobile deployment, \emph{Mobile-O-0.5B} components are converted using MLX~\cite{applemlx} and CoreML~\cite{applecoreml}. The language model runs in MLX Swift with 8-bit weights on GPU for efficient token decoding, while the vision encoder, DiT backbone, VAE decoder, and MCP are exported to Core ML in float32, keeping the total memory footprint below 2GB.

\begin{table}[t]
\centering
\caption{\textbf{Image understanding and generation performance comparison} on MacBook M2 Pro, Jetson Orin Nano, and iPhone for \emph{Mobile-O-0.5B}. Vision Enc. and TTFT denote understanding latency, while Latency indicates image generation latency.}
\label{tab:image_understanding_generation}
\resizebox{\columnwidth}{!}{
\begin{tabular}{lccc}
\toprule
\textbf{Model} & \textbf{Vision Enc. (ms)} & \textbf{TTFT (ms)} & \textbf{Latency (s)} \\
\midrule
\multicolumn{4}{c}{\textbf{MacBook M2 Pro}} \\
\midrule
Janus         & $783 \pm 244$  & $289 \pm 19$   & $201 \pm 15614$ \\
JanusFlow     & $1909 \pm 466$ & $935 \pm 152$  & $24 \pm 0.8$ \\
Show-O        & $699 \pm 107$  & $797 \pm 5$    & $47 \pm 0.2$ \\
\rowcolor{front-color}
\textbf{Mobile-O-0.5B (Ours)} & $\mathbf{56} \pm \mathbf{7}$ & $\mathbf{187} \pm \mathbf{31}$ & $\mathbf{4} \pm \mathbf{0.5}$ \\
\midrule
\multicolumn{4}{c}{\textbf{Jetson Orin Nano}} \\
\midrule
Janus         & $745 \pm 19$   & $749 \pm 19$   & $44 \pm 0.8$ \\
JanusFlow     & $741 \pm 27$   & $745 \pm 27$   & $22 \pm 0.1$ \\
Show-O        & $403 \pm 4$    & $720 \pm 14$   & $52 \pm 4$ \\
\rowcolor{front-color}
\textbf{Mobile-O-0.5B (Ours)} & $\mathbf{88} \pm \mathbf{7}$ & $\mathbf{488} \pm \mathbf{9}$ & $\mathbf{4} \pm \mathbf{0.6}$ \\
\midrule
\multicolumn{4}{c}{\textbf{iPhone 17 Pro}} \\
\midrule
\rowcolor{front-color}
\textbf{Mobile-O-0.5B (Ours)} & $\mathbf{102} \pm \mathbf{4}$ & $\mathbf{248} \pm \mathbf{10}$ & $\mathbf{3} \pm \mathbf{0.5}$ \\
\bottomrule
\end{tabular}
}
\end{table}

\section{Conclusion}
\label{sec:conclusion}
We introduce a unified vision–language–diffusion model, \emph{Mobile-O}, with a new quadruplets format for unified post-training and \emph{mobile conditioning projector} to achieve high-quality image understanding and text-to-image generation on edge devices. Experiments on MacBook M2 Pro, Jetson Orin Nano, and iPhone device show that \emph{Mobile-O} outperforms recent unified models in both latency and memory efficiency, while preserving visual fidelity and semantic accuracy. \emph{Mobile-O-0.5B} maintains a memory footprint below 2GB on iPhone within $\sim$3 seconds, making it practical for real-time on-device deployment.

\section{Acknowledgment}
\label{sec:Acknowledgment}
The computations were enabled by resources provided by NAISS at Alvis
partially funded by Swedish Research Council through grant agreement no. 2022-06725, LUMI
hosted by CSC (Finland) and LUMI consortium, and by Berzelius resource provided by the Knut and
Alice Wallenberg Foundation at the NSC.

{
    \small
    \bibliographystyle{ieeenat_fullname}
    \bibliography{main}
}

\clearpage
\maketitlesupplementary

\section{Mobile Conditioning Projector Depth}

The Mobile Conditioning Projector (MCP) aggregates features from multiple VLM layers to provide rich semantic conditioning for the diffusion model. Tab.~\ref{tab:mcp_ablation_2} investigates how the number of aggregated layers affects text-to-image generation quality on GenEval. Using a single layer yields 68.7\% accuracy, suggesting that features from one depth level provide insufficient semantic diversity for accurately capturing complex compositional prompts. Aggregating 2 layers with learnable fusion improves performance to 69.8\%, demonstrating the value of combining features from different network depths. The best performance (70.4\%) is achieved with 4 layers, striking an optimal balance between semantic richness and computational efficiency. Interestingly, further increasing to 8 layers slightly degrades performance to 70.2\%, indicating that excessive aggregation may introduce redundant or conflicting information that complicates the conditioning process. With four layers, it suggests that mid-depth VLM features capture the most relevant semantic abstractions for guiding compositional image generation, while avoiding the diminishing returns and increased computational cost associated with deeper aggregation.

{
\setlength{\textfloatsep}{1pt} 
\setlength{\intextsep}{1pt}    
\begin{table}[t]
\centering
\caption{
\textbf{Ablation on the number of layers for the Mobile Conditioning Projector (MCP).}
We systematically vary the number of VLM layers aggregated by MCP to condition the diffusion process. All configurations use the final design of MCP: learnable fusion, compression , and channel attention (CA).}
\label{tab:mcp_ablation_2}
\resizebox{\columnwidth}{!}{
\begin{tabular}{lccccc}
\toprule
\textbf{Proj.} & \textbf{\# Layers} & \textbf{Fusion} & \textbf{Compress} & \textbf{CA} & \textbf{Accuracy (\%)}  \\
\midrule
MCP  & 1 & - & \checkmark & \checkmark  & 68.7 \\
MCP & 2 & Learnable & \checkmark & \checkmark & 69.8\\
\rowcolor{front-color}
MCP & 4 & Learnable & \checkmark & \checkmark & 70.4\\
MCP & 8 & Learnable & \checkmark & \checkmark & 70.2\\
\bottomrule
\end{tabular}
}
\end{table}
}

\section{On-Device Mobile Deployment}

Fig.~\ref{fig:mobile_figure} demonstrates Mobile-O running natively on an iPhone 17 Pro, validating the practical feasibility of deploying unified models on consumer mobile devices. The implementation showcases both core capabilities within a chat-based interface: text-to-image generation produces a detailed Bengal tiger image from a complex compositional prompt in 3 seconds, while image-to-text generation provides rich visual descriptions analyzing scene composition, subject positioning, depth perception, and atmospheric qualities in 0.3 seconds for text token forward time. The chat-based interface enables seamless switching between understanding and generation tasks within a single unified model, showcasing practical mobile AI applications without cloud dependency, ensuring user privacy and enabling offline functionality—critical requirements for real-world mobile applications. This deployment validates our architectural optimizations for the design choices, including the Mobile Conditioning Projector, proving that an efficient yet effective unified model can maintain high-quality unified capabilities with less than 2GB of memory.

\begin{figure}
    \centering
    \includegraphics[width=\linewidth]{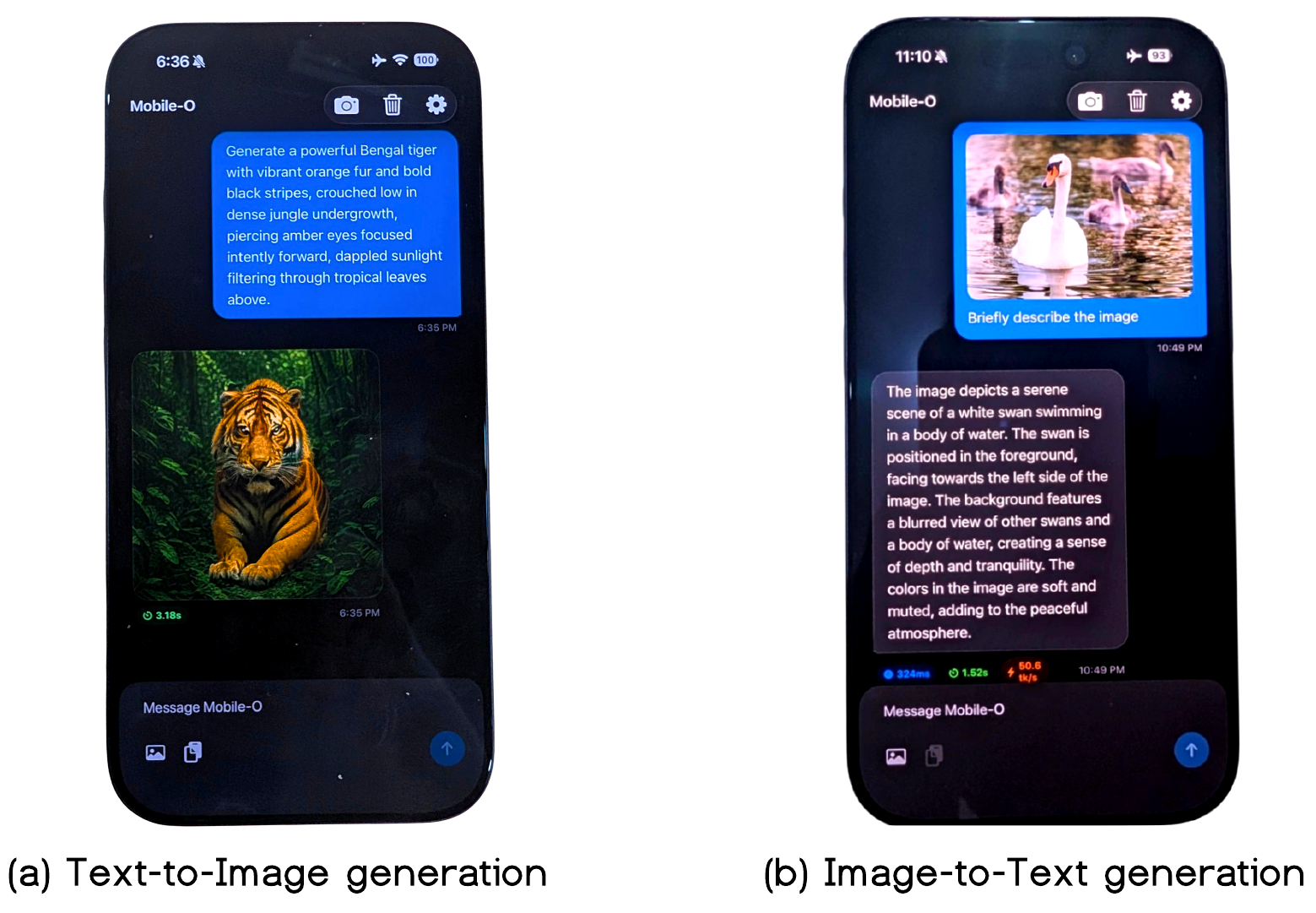}
    \caption{\textbf{Mobile-O running natively on iPhone 17 Pro.} We demonstrate real-world deployment of \textit{Mobile-O}'s unified capabilities on consumer hardware. (a) Text-to-image generation: Given a detailed prompt describing a Bengal tiger. (b) Image-to-text generation: Mobile-O provides detailed visual descriptions, analyzing composition and subject positioning}
    \label{fig:mobile_figure}
\end{figure}

\section{More Implementation Details}

All experiments are conducted on a single node with 8 NVIDIA A100 GPUs (80GB VRAM). We employ DeepSpeed ZeRO-3 during Stage 1 to efficiently handle the 9M training samples and large model parameters, then switch to ZeRO-1 for the last two stages, where smaller dataset sizes allow for reduced communication overhead. Mixed-precision training with BF16 throughout due to better numerical stability with transformer architectures. TF32 is enabled for matrix multiplications to leverage Ampere architecture acceleration. Images for understanding tasks undergo bicubic interpolation to 1024$\times$1024, while generation tasks use 512$\times$512.

We use LoRa with reduced rank (r=16) and $\alpha$=32 to prevent overfitting during unified training on the smaller 105K quadruplet dataset while still allowing fine-grained adaptation. All LoRA modules use a dropout of 0.1 for regularization. All stages use cosine annealing with minimum learning rate thresholds: \textbf{Stage 1:} LR decays from 2e-4 to 2e-6 over 50K steps with 2\% warmup (1,000 steps), allowing aggressive initial learning while maintaining stability in later training. \textbf{Stage 2:} LR decays from 2e-4 to 1e-6 with 5\% warmup, providing more gradual adaptation for the targeted fine-tuning phase. \textbf{Stage 3:} Reduced initial LR of 1e-4 (min: 1e-6) with 5\% warmup accommodates the unified training paradigm's increased complexity.

\begin{table*}[ht]
\centering
\resizebox{1.99\columnwidth}{!}{%
\setlength{\tabcolsep}{5pt}
\begin{tabular}{lccc}
\toprule
                 & \textit{Stage 1: Cross-Modal Alignment} & \textit{Stage 2: Supervised Fine-tuning} &  \textit{Stage 3: Unified Post-Training} \\
\midrule
Data Source      & JourneyDB (4M) + & BLIP3o-60K + & BLIP3o-60K + \\
                 & BLIP3o-Short (5M) & ShareGPT-4o (45K) & ShareGPT-4o (45K) \\
\midrule
Total Samples            & 9M                & 105K          & 105K \\
Format           & Prompt-image pairs & Prompt-image pairs & Quadruplet $\{p, \mathbf{x}_{\text{img}}, q, a\}$ \\
\midrule
Learning Rate    & 2e-4           & 2e-4         & 1e-4      \\
Batch Size       & 512  & 384 & 128 \\
LR Schedule      & cosine w/ min LR   & cosine w/ min LR & cosine w/ min LR     \\
Min LR           & 2e-6           & 1e-6         & 1e-6  \\
LR Warmup Ratio  & 0.02           & 0.05         & 0.05  \\
Optimizer        & AdamW ($\beta_2$=0.95) & AdamW ($\beta_2$=0.95) & AdamW ($\beta_2$=0.95)\\ 
Weight Decay     & 0.01           & 0.01         & 0.01 \\
\midrule
Epochs           & 5              & 20           & 7 \\
Training Time    & $\sim$3 days   & $\sim$15 hours & $\sim$5 hours \\
\midrule
Trainable Modules        & DiT + MCP      & DiT + MCP    & DiT + MCP + LLM + VE \\
Frozen Modules          & VE + LLM + VAE & VE + LLM + VAE & VAE \\

\bottomrule   
\end{tabular}
}
\caption{Three-stage training setup for Mobile-O. Stage 1 establishes cross-modal alignment using large-scale image-text pairs. Stage 2 performs targeted fine-tuning to address weaknesses in complex gestures, common objects, and landmarks. Stage 3 introduces unified multimodal post-training with quadruplet samples $\{p, \mathbf{x}_{\text{img}}, q, a\}$ for joint understanding and generation. All experiments were conducted on 8$\times$A100 GPUs.}
\label{tab:hyp-scaled}
\end{table*}

\section{More Image-to-Text Qualitative Results}

\begin{figure}
    \centering
    \includegraphics[width=\linewidth]{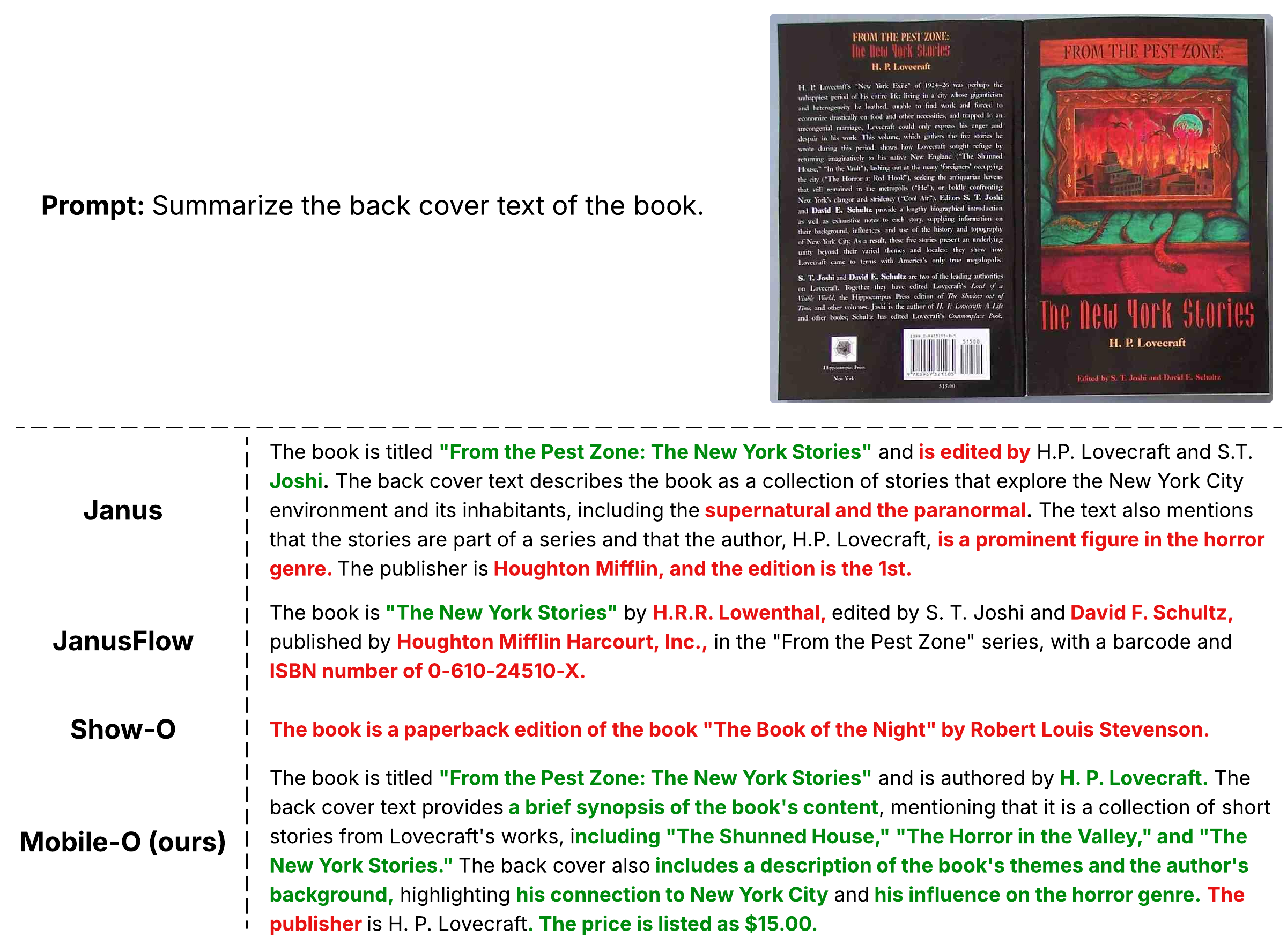}
    \caption{\textbf{Qualitative comparison on dense text understanding and information extraction.} We evaluate \textit{Mobile-O} against other models on a challenging OCR and comprehension task requiring the model to read, parse, and summarize the back cover text of a book. Green text indicates correctly extracted information, while red indicates hallucinations or errors. \textit{Mobile-O} demonstrates superior performance in accurately extracting key bibliographic details, including the correct title, author, editors, and price information from the densely-packed text on the book cover.}

    \label{fig:suppl_qualitative_understanding_dense}
\end{figure}

\begin{figure}[t]
    \centering
    \includegraphics[width=1.0\linewidth]{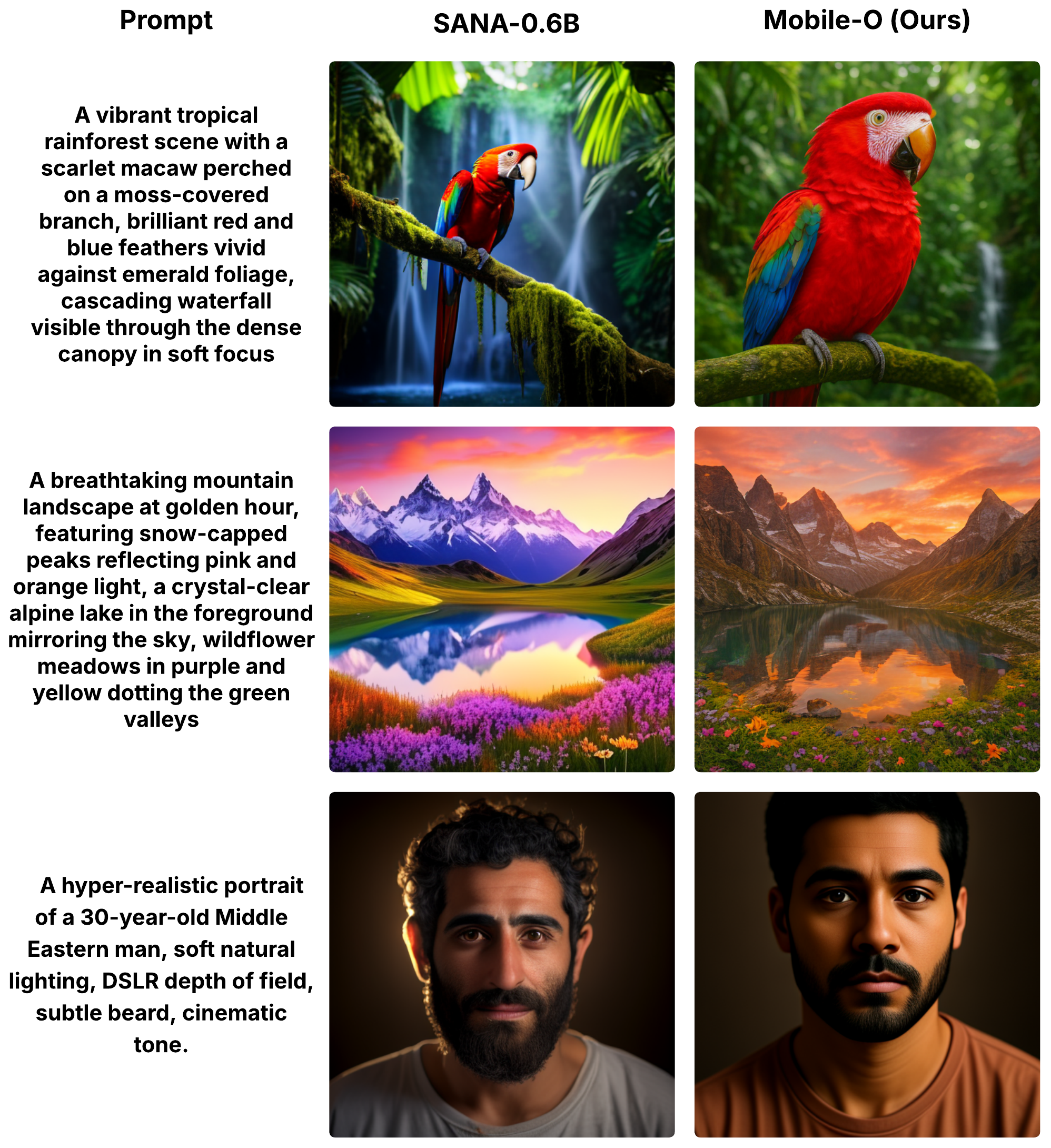}
    \caption{\textbf{Qualitative comparison with SANA-0.6B on text-to-image generation.} We compare \textit{Mobile-O} (1.6B total parameters) against SANA-0.6B (2.6B total parameters), our generation baseline, on challenging prompts requiring photorealistic rendering, complex lighting, and fine-grained details. \textit{Mobile-O} demonstrates competitive or superior visual quality across diverse scenarios, including wildlife photography, landscape composition, and portrait rendering. Best viewed zoomed in.}
    \label{fig:sana_comparison}
\end{figure}

\begin{figure*}
    \centering
    \includegraphics[width=\linewidth]{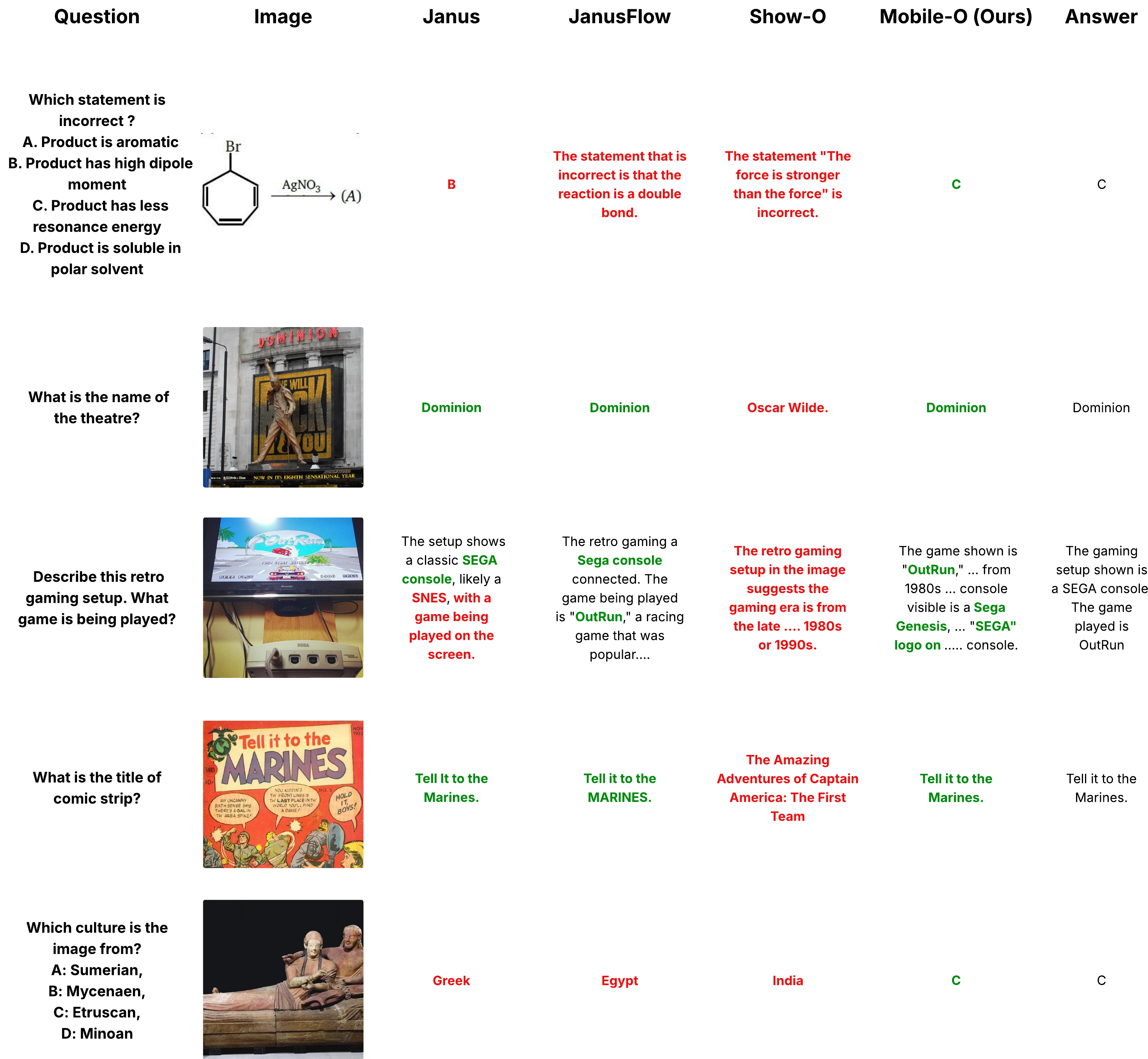}
    \caption{\textbf{Qualitative comparison of image-to-text across unified models below 2B.} \textit{Mobile-O} is compared against Janus~\cite{janus}, JanusFlow~\cite{ma2025janusflow}, and Show-O~\cite{showo} on diverse visual question answering tasks, including scientific reasoning, OCR, object recognition, and cultural knowledge. Green indicates correct answers, red indicates errors. \textit{Mobile-O} demonstrates competitive visual understanding, despite its mobile-optimized architecture, correctly answering complex questions that require fine-grained visual analysis and domain knowledge.}

    \label{fig:suppl_qualitative_understanding}
\end{figure*}

\begin{figure*}
    \centering
    \includegraphics[width=\linewidth]{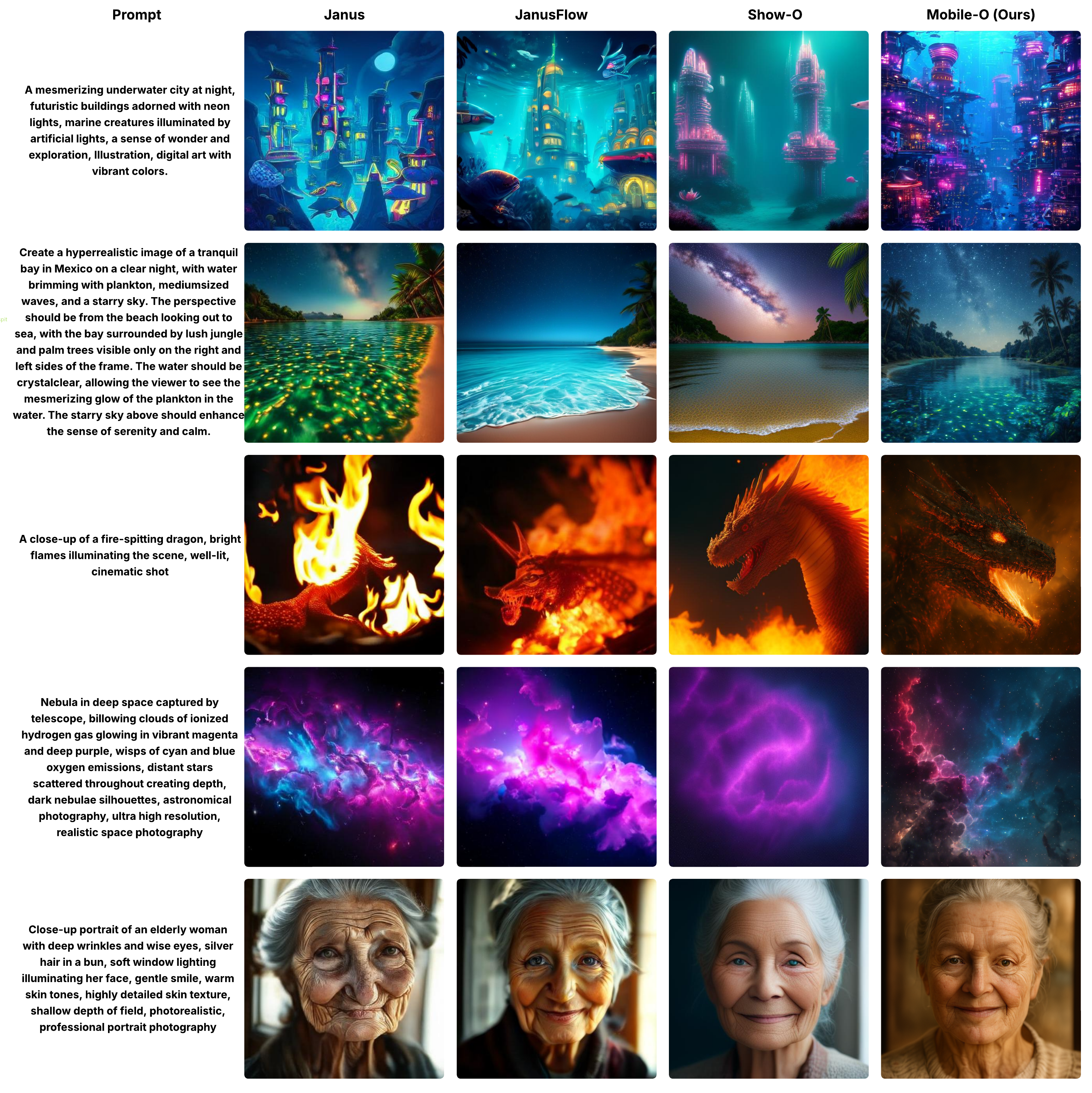}
        \caption{\textbf{Qualitative comparison of text-to-image generation across unified models below 2B.} \textit{Mobile-O} is compared against Janus~\cite{janus}, JanusFlow~\cite{ma2025janusflow}, and Show-O~\cite{showo} on challenging prompts spanning fantasy, photorealism, and scientific visualization. Despite its mobile-optimized architecture, \textit{Mobile-O} maintains competitive visual quality and prompt adherence. Best viewed zoomed in.}

    \label{fig:qualitative_comparison_suppl}
\end{figure*}

\begin{figure*}
    \centering
    \includegraphics[width=\linewidth]{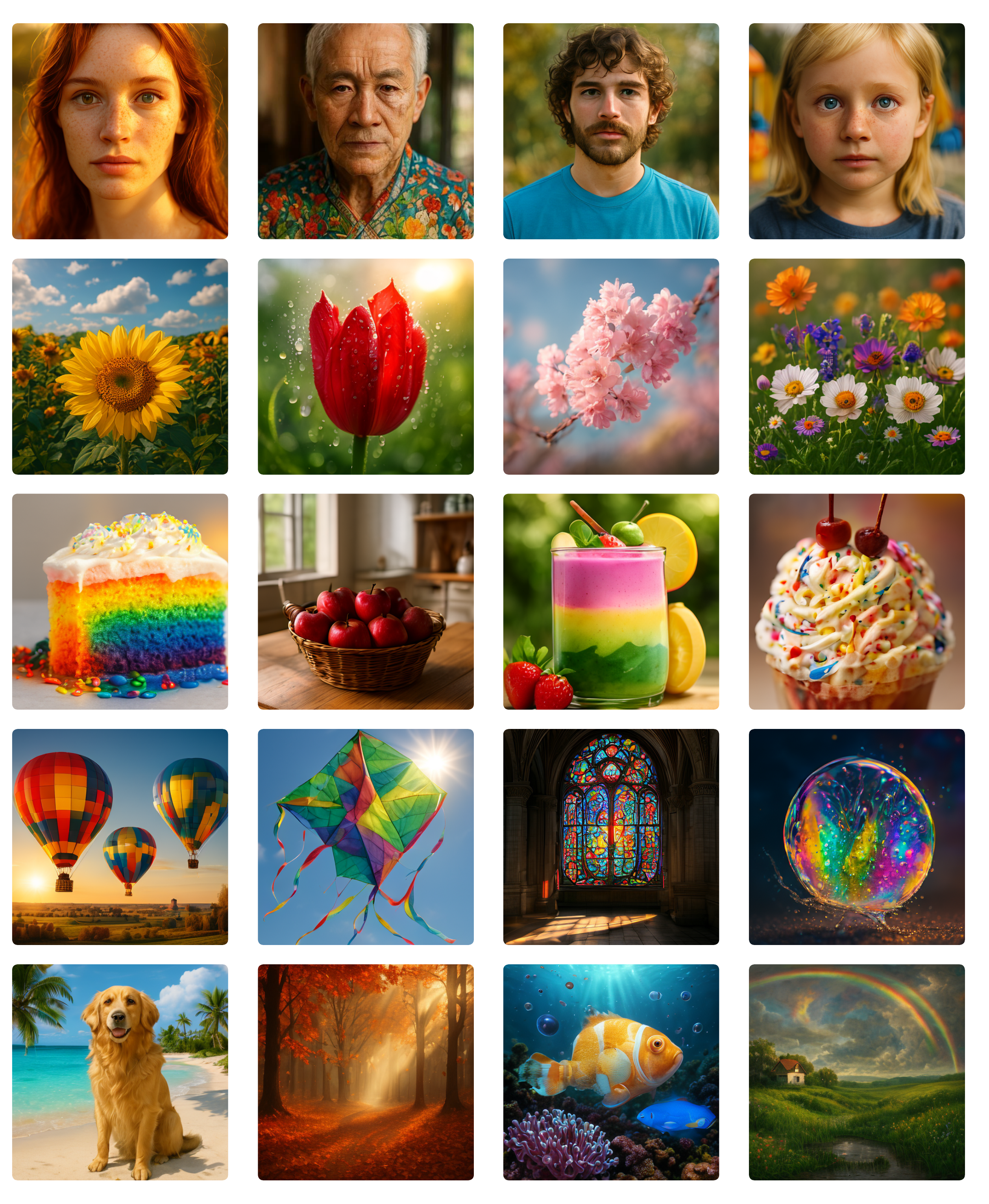}
    \caption{\textbf{Additional Text-to-Image generation examples of \textit{Mobile-O}.} Best viewed zoomed in.}
    \label{fig:qualitative_results_suppl}
\end{figure*}

Fig.~\ref{fig:suppl_qualitative_understanding_dense} evaluates the models' ability to perform dense text understanding and information extraction from real-world imagery. The task requires reading small, low-contrast text from a book's back cover and summarizing its bibliographic information—a challenging scenario combining OCR, reading comprehension, and structured information extraction. \textit{Mobile-O} accurately identifies the book as "From the Pest Zone: The New York Stories" authored by H.P. Lovecraft, correctly extracts the editors' names (S.T. Joshi and David E. Schultz), identifies specific story titles mentioned in the synopsis, and even captures the price 15.00 USD. In contrast, competing models exhibit significant hallucinations and misidentify the book title, authors, and fail to display the price. These results validate \textit{Mobile-O}'s robust text understanding capabilities even in challenging real-world conditions with dense text, complex layouts, and varying contrast levels.

Fig.~\ref{fig:suppl_qualitative_understanding} presents a comprehensive qualitative evaluation of visual understanding capabilities across unified vision-language models on diverse question-answering tasks. The comparison spans multiple cognitive domains including scientific reasoning (organic chemistry reaction analysis), optical character recognition with challenging perspectives and lighting conditions (theater signage reading), fine-grained object recognition requiring specific domain knowledge (retro gaming console and software identification), text extraction from stylized fonts (comic book titles), and cultural artifact classification (ancient civilization identification) from MMMU~\cite{yue2024mmmu}, ChartQA~\cite{masry2022chartqa}, and TextVQA~\cite{singh2019textvqa}. These results validate that \textit{Mobile-O}'s mobile-optimized architecture preserves robust visual understanding capabilities, demonstrating that aggressive model compression need not compromise the ability to accurately interpret and reason about diverse visual information.

\section{Comparison with Generation-Only Baseline}

Fig.~\ref{fig:sana_comparison} compares \textit{Mobile-O} against SANA-0.6B, the generation component that serves as our baseline architecture. Despite \textit{Mobile-O} having 1.6B total parameters compared to SANA-0.6B's 2.6B parameters (38\% reduction), \textit{Mobile-O} achieves competitive or superior generation quality across diverse prompts. In the rainforest scene, \textit{Mobile-O} produces sharper feather details and a more natural background compared to SANA's slightly oversaturated rendering. For the mountain landscape, \textit{Mobile-O} captures more realistic geological textures and natural color grading, while SANA exhibits somewhat exaggerated saturation in the foreground flowers. The portrait comparison reveals \textit{Mobile-O}'s superior handling of skin tones and facial features with more natural lighting and realistic depth of field. \textit{Mobile-O} achieves these results while simultaneously supporting visual understanding tasks within the model.

\section{More Text-to-Image Qualitative Results}

Fig.~\ref{fig:qualitative_comparison_suppl} presents a comprehensive qualitative comparison between \textit{Mobile-O} and recent unified models across diverse and challenging prompts. The comparison includes Janus~\cite{janus}, JanusFlow~\cite{ma2025janusflow}, and Show-O~\cite{showo}, evaluating generation quality on prompts ranging from fantastical scenes (underwater cities, fire-breathing dragons) to photorealistic scenarios (bio-luminescent bays, space nebulae, portrait photography). \textit{Mobile-O} demonstrates competitive visual quality while maintaining significantly lower computational requirements suitable for mobile deployment. Notably, \textit{Mobile-O} excels at rendering fine details and maintaining prompt adherence across complex compositional scenarios, such as the intricate architectural details in the underwater city scene and the nuanced lighting in the portrait photography example. While competing models occasionally produce visually striking results, \textit{Mobile-O} achieves a favorable balance between generation quality, prompt fidelity, and computational efficiency. The nebula scene particularly highlights \textit{Mobile-O}'s ability to capture subtle color gradations and spatial depth, while the elderly woman portrait demonstrates proficient handling of photorealistic skin textures and natural lighting.

Fig.~\ref{fig:qualitative_results_suppl} showcases\textit{ Mobile-O}'s text-to-image generation capabilities across diverse categories, including photorealistic portraits, macro nature photography, food imagery, and creative scenes with complex lighting effects. The model demonstrates proficiency in rendering fine details (facial features, textures), managing challenging optical effects (bokeh, volumetric lighting, caustics), and maintaining color accuracy across varied subjects. These results validate \textit{Mobile-O}'s versatility in generating high-quality imagery across different styles and compositional complexities while operating within mobile computational constraints. The prompts used in Fig.~\ref{fig:qualitative_results_suppl} are provided in Tab.~\ref{tab:qualitative_prompts}.

\label{sec:suppl_material}

\section{Limitations}
\textit{Mobile-O} currently reuses the same lightweight LLM from the unified VLM as its text encoder, rather than employing a dedicated standalone language model optimized solely for textual understanding. This design choice significantly reduces memory footprint and allows on-device deployment, but it may limit the expressiveness and depth of text representations compared to approaches that use larger text-only models. For instance, SANA~\cite{sana} adopts Gemma-2B-it~\cite{gemma2_2024} as a dedicated text encoder, benefiting from a more powerful linguistic backbone that can yield better alignment.

However, integrating such a model into \textit{Mobile-O} is currently impractical for on-device deployment. A 2B-parameter model in FP16 requires approximately 4.0 GB just for the weights alone, excluding memory for activations, attention caches, and runtime overhead, which typically increases total memory requirements by several additional GBs. This exceeds the memory constraints of most mobile and resource-limited edge devices, where efficiency and low latency are core deployment objectives.

\begin{table*}[t]
\centering
\caption{Text-to-image generation prompts used for visualization.}
\label{tab:qualitative_prompts}
\renewcommand{\arraystretch}{1.2}
\begin{tabular}{|l|p{0.88\textwidth}|}
\hline
\rowcolor{black!60}
\multicolumn{2}{|c|}{\textbf{\textcolor{white}{Detailed Prompts for Image Generation in Fig.~\ref{fig:qualitative_results_suppl} --- $(i,j)$ denotes the image at row $i$, column $j$}}} \\
\hline
\rowcolor{gray!15}
\textbf{Row 1} & \textbf{Portrait Photography} \\
\hdashline[2pt/2pt]
(1,1) & Young woman with freckles, green eyes with detailed iris and catchlight, natural skin texture, flowing red hair catching golden hour sunlight, warm vibrant lighting, gentle expression, shallow depth of field \\[1mm]
(1,2) & Elderly Asian man with silver hair and peaceful expression, kind brown eyes, colorful traditional clothing, bright natural window lighting, detailed skin texture \\[1mm]
(1,3) & Young man with curly hair and beard, bright turquoise shirt, outdoor lighting \\[1mm]
(1,4) & Child with bright blue eyes and blonde hair, curious expression, rosy cheeks, individual eyelashes visible, natural freckles, bright daylight from side creating dimension, colorful playground in blurred background \\[2mm]
\hline
\rowcolor{gray!15}
\textbf{Row 2} & \textbf{Flowers \& Garden Scenes} \\
\hdashline[2pt/2pt]
(2,1) & Sunflower field with detailed centers, yellow petals with texture, bees visiting flowers, blue sky with white clouds, warm summer sunlight, cheerful atmosphere, depth of field \\[1mm]
(2,2) & Single red tulip with dewdrops on petals, water droplets reflecting light, delicate petal veins, green stem, bright spring morning light, shallow depth of field \\[1mm]
(2,3) & Cherry blossom branch in full bloom, pink petals with detailed stamens, some petals floating in air, bright blue sky background, spring sunshine creating glow, intricate branch structure \\[1mm]
(2,4) & Cluster of wildflowers, purple lupines and orange California poppies with detailed petal texture, yellow daisy centers, white Queen Anne's lace intricate patterns, bright midday sun, morning dew \\[2mm]
\hline
\rowcolor{gray!15}
\textbf{Row 3} & \textbf{Food \& Culinary Compositions} \\
\hdashline[2pt/2pt]
(3,1) & Rainbow cake slice showing colorful layers, white frosting, sprinkles, bright studio lighting \\[1mm]
(3,2) & Basket with red apples on wooden kitchen table, bright natural window lighting \\[1mm]
(3,3) & Colorful layered smoothie in clear glass, vibrant pink strawberry, purple acai, yellow mango, green spinach layers, fresh fruit garnish, straw, bright natural lighting \\[1mm]
(3,4) & Ice cream sundae with rainbow sprinkles, individual sprinkle shapes and colors sharp, melting ice cream texture, whipped cream peaks, glossy cherry, colorful syrup drizzle, bright studio lighting \\[2mm]
\hline
\rowcolor{gray!15}
\textbf{Row 4} & \textbf{Colorful Objects \& Items} \\
\hdashline[2pt/2pt]
(4,1) & Colorful hot air balloons in mid-air, red, yellow, and blue, wicker baskets, golden morning sunlight, blue sky, countryside below \\[1mm]
(4,2) & Colorful kite with rainbow geometric patterns, ribbon tails, bright blue sky, sunlight \\[1mm]
(4,3) & Ornate stained glass window with intricate patterns in reds, blues, yellows, greens, bright sunlight streaming through, colorful light on floor \\[1mm]
(4,4) & Soap bubble bursting in mid-air, water droplets spraying outward, iridescent rainbow colors on fragmenting bubble surface, dynamic motion, dramatic lighting \\[2mm]
\hline
\rowcolor{gray!15}
\textbf{Row 5} & \textbf{Natural Landscapes \& Outdoor Scenes} \\
\hdashline[2pt/2pt]
(5,1) & Golden retriever dog sitting on tropical beach, turquoise water, white sand, bright blue sky, sunny day \\[1mm]
(5,2) & Autumn forest path with vibrant orange, red, and yellow fall foliage, sun rays piercing through misty air, leaves falling mid-flight, dramatic golden light beams, warm glowing atmosphere \\[1mm]
(5,3) & Orange and white clownfish among pink sea anemone tentacles with texture, vibrant purple and yellow corals with polyp detail, blue tang fish nearby, bright sunlight filtering through water creating god rays, bubbles rising \\
(5,4) & Landscape after rain with vibrant double rainbow arching across sky, green rolling hills with visible grass texture, wildflowers, small farmhouse, sunlight breaking through dramatic clouds, puddle in foreground reflecting rainbow, wet grass sparkling \\
\hline
\end{tabular}
\end{table*}

\end{document}